\documentclass{bmvc2k}

\usepackage{adjustbox}
\usepackage{afterpage}
\usepackage{algorithm}
\usepackage[noend]{algpseudocode}
\usepackage{amsmath}
\usepackage{amssymb}

\usepackage{makecell}
\usepackage{multirow}

\usepackage{graphicx}
\usepackage{tikz}
\usepackage{tikzscale}
\usetikzlibrary{positioning}

\makeatletter
\algnewcommand{\LineComment}[1]{\Statex \hspace{-13pt} \ALG@thistlm \(\triangleright\) #1}
\makeatother

\DeclareMathOperator*{\argmax}{arg\,max}

\newcommand{\zeropar}[1]{
    \vspace{-27pt}
    \noindent
    \refstepcounter{subsection}
    \paragraph{\the\numexpr\value{section}.\the\numexpr\value{subsection}\,\,#1:}
}

\title{Explaining Multi-modal Large Language Models by Analyzing their Vision Perception}

\addauthor{Loris Giulivi}{}{1}
\addauthor{Giacomo Boracchi}{}{1}

\addinstitution{
 Politecnico di Milano 
}

\runninghead{Giulivi, Boracchi}{Explaining MLLMs}

\begin{document}

\maketitle
\vspace{-6pt}
\begin{abstract}
Multi-modal Large Language Models (MLLMs) have demonstrated remarkable capabilities in understanding and generating content across various modalities, such as images and text. However, their interpretability remains a challenge, hindering their adoption in critical applications. This research proposes a novel approach to enhance the interpretability of MLLMs by focusing on the image embedding component. We combine an open-world localization model with a MLLM, thus creating a new architecture able to simultaneously produce text and object localization outputs from the same vision embedding.
The proposed architecture greatly promotes interpretability, enabling us to design a novel saliency map to explain any output token, to identify model hallucinations, and to assess model biases through semantic adversarial perturbations.
\end{abstract}

\vspace{-9pt}
\section{Introduction}
\label{sec:intro}
Since the advent of Chat-GPT, a large language model (LLM) revolution has taken the Machine Learning (ML) community by storm.
More recently, Multi-modal Large Language Models (MLLMs), able to reason on inputs composed of both images and text \cite{mllmsurvey}, have shown even more impressive results on many Computer Vision (CV) problems.
MLLMs such as Flamingo \cite{flamingo}, LLaVa \cite{llava}, and GPT-4 \cite{gpt4} are now able to solve a plethora of language and vision tasks with a level of accuracy that was unthinkable just a few years ago.

Consequently, the research community has focused on improving the performance of MLLMs, rather than assessing their interpretability or developing explanations. Indeed, the most popular techniques to explain transformers, Attention Visualization \cite{vit} and Attention Rollout \cite{attentionrollout}, predate the introduction of MLLMs by years.
This issue is further magnified by the predisposition of MLLMs towards biases~\cite{mllmbias} and hallucinations~\cite{mllmhallucinations}.

\begin{figure}
    \centering
    \resizebox{0.9\linewidth}{!}{
        \centering
        \begin{tikzpicture}
            \node[anchor=south west, inner sep=0] (image) at (0,0) {\includegraphics[width=0.9\textwidth]{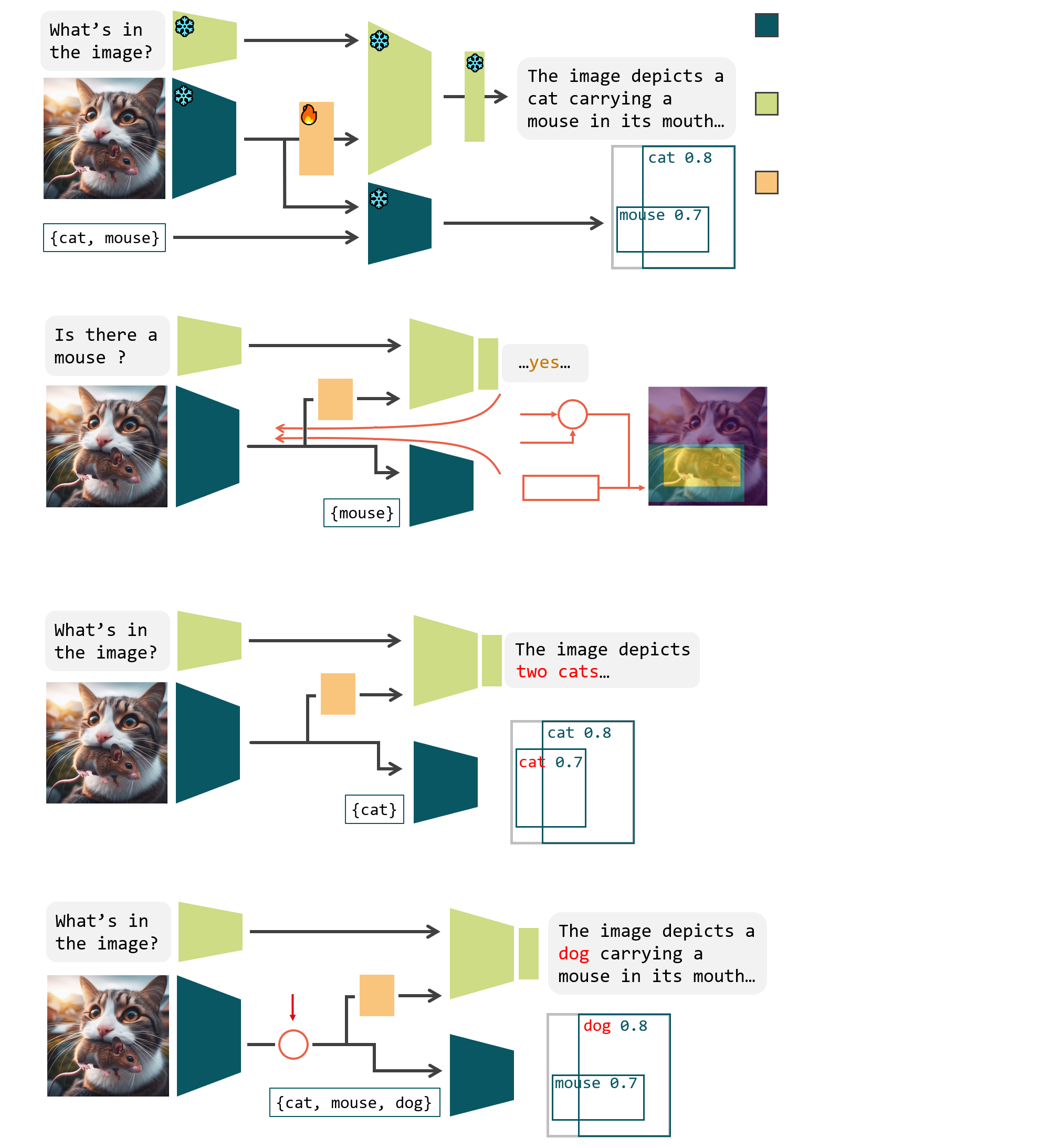}};
            \node[above right, text=black, rotate=90] at (-0.5, 10.2) {Architecture};

            \node[above right, text=black, rotate=0] at (0.0, 12.1) {$\mathbf{p}$};
            \node[above right, text=black, rotate=0] at (0.0, 11.0) {$\mathbf{x}$};
            \node[above right, text=black, rotate=0] at (0.0, 9.8) {$\mathbf{Q}$};
            
            \node[above right, text=black, rotate=0] at (2.0, 12.0) {\small $\mathcal{T}$};
            \node[above right, text=white, rotate=0] at (1.86, 11.0) {\tiny $\mathcal{E}^{OWL}$};
            \node[above right, text=black, rotate=0] at (3.97, 11.4) {\small LLM};
            \node[above right, text=white, rotate=0] at (4.15, 10.0) {\small $\mathcal{H}$};
            \node[above right, text=black, rotate=90] at (5.45, 11.1) {\tiny $decode$};
            \node[above right, text=black, rotate=0] at (3.25, 10.85) {\small $W$};

            \node[above right, text=black, rotate=0] at (2.7, 12.25) {\small $\mathbf{t}_i$};
            \node[above right, text=black, rotate=0] at (2.6, 11.1) {\tiny $\mathbf{t}^{OWL}_i$};
            \node[above right, text=black, rotate=0] at (3.64, 11.2) {\tiny $\mathbf{t}^{W}_i$};
            \node[above right, text=black, rotate=0] at (4.75, 11.6) {\small $\hat{\mathbf{l}}_i$};
            \node[above right, text=black, rotate=0] at (5.33, 11.6) {\small $\hat{\mathbf{t}}_i$};
            \node[above right, text=black, rotate=0] at (4.75, 10.3) {\fontsize{7}{9}\selectfont $\{\mathbf{z}_i,x_i,y_i,w_i,h_i\}$};

            \node[above right, text=black, rotate=0] at (8.6, 11.75) {\small \parbox{85pt}{Frozen modules taken from OWL-ViT~\cite{owlvit-t}.}};
            \node[above right, text=black, rotate=0] at (8.6, 10.9) {\small \parbox{85pt}{Frozen modules taken from LLaVa~\cite{llava}.}};
            \node[above right, text=black, rotate=0] at (8.6, 10.4) {\small \parbox{90pt}{Trained alignment layer.}};
            
            \draw[black, thick] (0,9.5) -- (image.south east |- 0,9.5);

            \node[above right, text=black, rotate=90] at (-0.4, 6.9) {Saliency map};
            
            \node[above right, text=black, rotate=0] at (2.0, 8.6) {\small $\mathcal{T}$};
            \node[above right, text=black, rotate=0] at (3.45, 8.05) {\small $W$};
            \node[above right, text=white, rotate=0] at (1.9, 7.55) {\tiny $\mathcal{E}^{OWL}$};
            \node[above right, text=black, rotate=0] at (2.55, 7.70) {\small $\mathbf{t}_i^{OWL}$};
            \node[above right, text=black, rotate=0] at (4.44, 8.45) {\small LLM};
            \node[above right, text=white, rotate=0] at (4.6, 7.1) {\small $\mathcal{H}$};
            
            \node[above right, text=black, rotate=0] at (5.3, 7.85) {\small $\nabla$};
            \node[above right, text=black, rotate=0] at (5.3, 7.50) {\small $\nabla$};
            \node[above right, text=black, rotate=0] at (6.12, 7.85) {\small $\phi$};
            
            \node[above right, text=black, rotate=0] at (5.35, 7.09) {\fontsize{7}{9}\selectfont $\mathbf{z}$};
            \node[above right, text=black, rotate=0] at (5.75, 7.09) {\fontsize{6}{9}\selectfont $x,y,w,h$};

            \node[above right, text=black, rotate=0] at (8.6, 6.5) {\small \parbox{90pt}{We compute 
            cosine similarity $\phi(\frac{\partial z_m}{\partial\mathcal{E}^{OWL}}, \frac{\partial\hat{\textbf{l}}_n}{\partial\mathcal{E}^{OWL}})$ to identify relevance between output tokens and output bounding boxes, thus creating a saliency map.}};
            
            \draw[black, thick] (0,6.5) -- (image.south east |- 0,6.5);

            \node[above right, text=black, rotate=90] at (-0.5, 3.6) {Hallucinations};
            
            \node[above right, text=black, rotate=0] at (2.0, 5.4) {\small $\mathcal{T}$};
            \node[above right, text=black, rotate=0] at (3.48, 4.8) {\small $W$};
            \node[above right, text=white, rotate=0] at (1.9, 4.25) {\tiny $\mathcal{E}^{OWL}$};
            \node[above right, text=black, rotate=0] at (2.6, 4.45) {\small $\mathbf{t}_i^{OWL}$};
            \node[above right, text=black, rotate=0] at (4.46, 5.2) {\small LLM};
            \node[above right, text=white, rotate=0] at (4.62, 3.85) {\small $\mathcal{H}$};

            \node[above right, text=black, rotate=0] at (8.6, 3.5) {\small \parbox{90pt}{The LLM and the detection model share the vision representation of the image $\mathbf{t}_i^{OWL}$, thus, errors are reflected on both outputs, enabling to visualize hallucinations.}};
            
            \draw[black, thick] (0,3.2) -- (image.south east |- 0,3.2);

            \node[above right, text=black, rotate=90] at (-0.5, 0.6) {Bias detection};
            
            \node[above right, text=black, rotate=0] at (2.0, 2.15) {\small $\mathcal{T}$};
            \node[above right, text=black, rotate=0] at (3.90, 1.45) {\small $W$};
            \node[above right, text=white, rotate=0] at (1.9, 1.0) {\tiny $\mathcal{E}^{OWL}$};
            \node[above right, text=black, rotate=0] at (4.89, 1.95) {\small LLM};
            \node[above right, text=white, rotate=0] at (5.03, 0.60) {\small $\mathcal{H}$};
            
            \node[above right, text=black, rotate=0] at (2.56, 1.1) {\tiny $\mathbf{t}_i^{OWL}$};
            \node[above right, text=black, rotate=0] at (3.28, 1.1) {\tiny $\mathbf{t}_i^{ADV}$};
            \node[above right, text=red, rotate=0] at (2.7, 1.7) {\fontsize{5}{6}\selectfont \parbox{30pt}{$\mathbf{C}^+=\{\text{\texttt{dog}}\}$ \\ $\mathbf{C}^-=\{\text{\texttt{cat}}\}$}};
            \node[above right, text=red, rotate=0] at (3.0, 0.96) {\fontsize{5}{6}\selectfont $+$\fontsize{6}{6}\selectfont$\mathbf{e}$};

            \node[above right, text=black, rotate=0] at (8.6, -0.2) {\small \parbox{90pt}{We compute perturbation $\mathbf{e}$ and add it to $\mathbf{t}_i^{OWL}$ such that the visual semantics are changed, and visualize its effects on the two outputs. We use this to detect and measure model bias.}};
            
            
            
        \end{tikzpicture}
    }
    \vspace{-10pt}
    \caption{Overview of the proposed architecture and its uses for interpretability.}
    \vspace{-15pt}
    \label{fig:teaser}
\end{figure}

These aspects highlight the urgent need for MLLM explanations. 
To tackle this issue, we present a joint open-world localization (OWL-ViT \cite{owlvit-t}) and MLLM (LLaVa \cite{llava}) model (Figure \ref{fig:teaser}), allowing for simultaneous extraction of text ($\mathbf{O}^{MLLM}$) and bounding boxes ($\mathbf{O}^{OWL}$) from the same vision embedding $\mathbf{t}_i^{OWL}$. 
In the proposed architecture, the detection output acts as a compact representation of how the MLLM interprets the image's semantics, displaying the objects that are perceived in the image. 
We exploit this property to \emph{detect and visualize model hallucinations}.
Moreover, by analyzing the gradients of $\mathbf{O}^{MLLM}$ and $\mathbf{O}^{OWL}$ with respect to the embedding $\mathbf{t}_i^{OWL}$, we \emph{develop a novel saliency map to explain the outputs of the MLLM}.
Lastly, enabled by the tight link between the two outputs, we design adversarial perturbations to $\mathbf{O}^{OWL}$ that reflect a semantic change to the shared embedding $\mathbf{t}_i^{OWL}$ and thus to $\mathbf{O}^{MLLM}$, and exploit these perturbations to \emph{assess and measure MLLM biases}.


To the best of our knowledge, our work is the first to enable explanations for the Vision Transformer (ViT) component of a MLLM.
Indeed, previous explanations for ViTs \cite{transformer,attentionrollout} are only able to explain the model's attention in relation to a training class, and cannot be applied to arbitrary token outputs of a downstream LLM.
Furthermore, we are the first to employ adversarial perturbations to the purposes of explaining a MLLM, as previous literature on the subject primarily focuses on deceiving models or defending from attacks~\cite{adversarialsurvey}.
We validate our saliency map by means of a user study, demonstrating that the proposed explanation identifies regions in the image that are relevant to the explained token. Moreover, we demonstrate that the very recent MLLM employed in this work~\cite{llava} is prone to biases and hallucinations. 
We make our code publicly available at the project's~\href{https://github.com/loris2222/ExplainingMLLMs}{github page}.


\vspace{-7pt}
\section{Background}
\vspace{-3pt}
\label{sec:background}
We now discuss relevant literature on the subjects of open-world object localization (OWL) and Multi-modal Large Language Models (MLLMs), on which we base our model, and on saliency maps, which are at the core of one of our proposed explanations.



\begin{figure}[t]
    \centering
    \resizebox{0.9\linewidth}{!}{
        \includegraphics{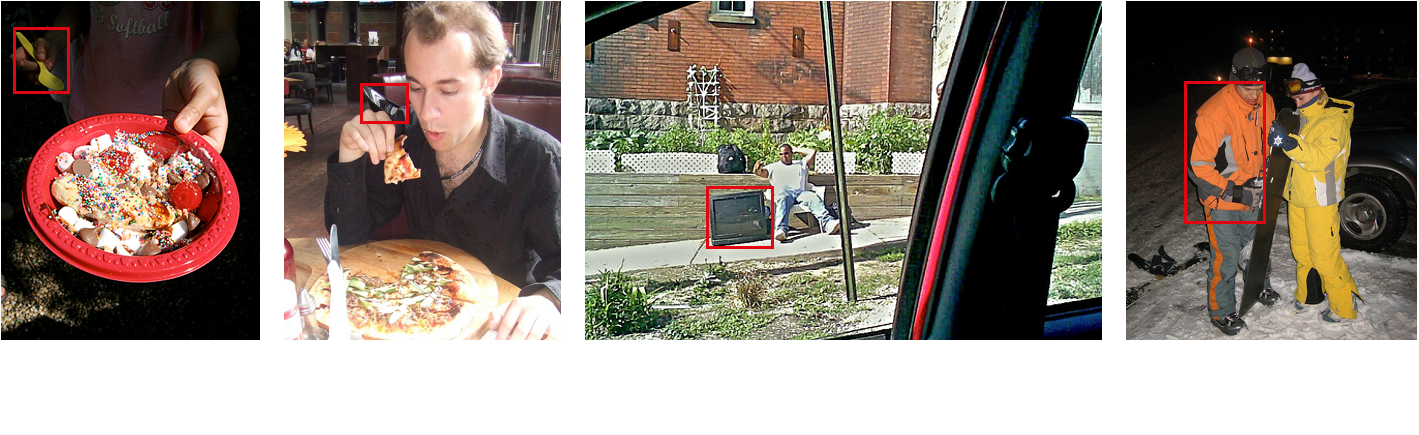}
    }
    \put(-344,60){\fontsize{8}{10}\selectfont \rotatebox[]{90}{$\mathbf{x}, \mathbf{O}^{OWL}$}}
    \put(-344,7){\fontsize{8}{10}\selectfont \rotatebox[]{90}{$\mathbf{O}^{MLLM}$}}
    \put(-327,95){\fontsize{6}{8}\selectfont \color{red} \texttt{knife (q=0.57)}}
    \put(-330,8){\fontsize{6}{8}\selectfont \parbox{60pt}{\textit{USER: Is there a knife in the image? \\ MLLM: Yes.}}}
    \put(-262,83){\fontsize{5}{7}\selectfont \color{red} \texttt{cell phone (q=0.46)}}
    \put(-264,8){\fontsize{6}{8}\selectfont \parbox{60pt}{\textit{USER: Is there a cell phone in the image? \\ MLLM: Yes.}}}
    \put(-166,59){\fontsize{6}{8}\selectfont \color{red} \texttt{suitcase (q=0.53)}}
    \put(-194,12){\fontsize{6}{8}\selectfont \parbox{100pt}{\textit{USER: Is there a suitcase in the image? \\ MLLM: Yes.}}}
    \put(-65,84){\fontsize{6}{8}\selectfont \color{red} \texttt{backpack (q=0.51)}}
    \put(-68,8){\fontsize{6}{8}\selectfont \parbox{60pt}{\textit{USER: Is there a backpack in the image? \\ MLLM: Yes.}}}
    
    \vspace{-5pt}
    \caption{Example images that lead to hallucinations. The error is reflected both in the language output ($\mathbf{O}^{MLLM}$) and in the detection output ($\mathbf{O}^{OWL}$).}
    \vspace{-5pt}
    \label{fig:hallucination}
\end{figure}

\begin{figure}[t]
    \centering
    \resizebox{0.95\linewidth}{!}{
        \includegraphics{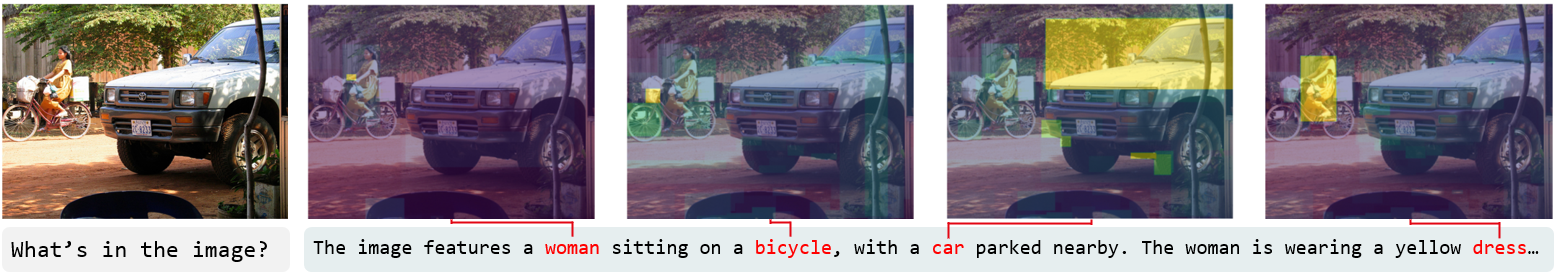}
    }
    \vspace{-10pt}
    \caption{Example GA saliency maps for different objects in one MLLM output.}
    \vspace{-40pt}
    \label{fig:saliency}
\end{figure}

\zeropar{Multi-modal Large Language Models}\label{sec:background-mllm}
MLLMs are models that can reason on both image and text modalities, returning a text output.
Formally, given a text prompt $\mathbf{p}$, its tokenized version $\mathcal{T}(\mathbf{p}) = \textbf{t}_1, \dots, \textbf{t}_n$, $\textbf{t}\in\mathbb{R}^d$, and an image $\mathbf{x}$, a MLLM outputs a logit sequence $\hat{\textbf{l}}_1, \dots, \hat{\textbf{l}}_n$, $\textbf{l}\in\mathbb{R}^{dict}$ with the same length as the input, which is then used to predict a token sequence $\hat{\textbf{t}}_1, \dots, \hat{\textbf{t}}_n$. 
Since the main application of MLLMs is chat-bots, these models are typically trained for \textit{causal language modeling}, that is 
$(\hat{\textbf{t}}_1, \dots, \hat{\textbf{t}}_n) \doteq (\textbf{t}_2, \dots, \textbf{t}_{n+1})$.


To include vision information in the language model, two main approaches have been proposed. The first consists in adding cross attention layers between image and text tokens~\cite{flamingo}, while the second relies on treating image tokens as part of the input sequence~\cite{llava,gpt4}. In both cases, a transformer network is employed to convert a raster image into a sequence of tokens that can be processed by the language model component of the MLLM.
In our work, we focus on MLLMs that treat vision tokens as part of the input sequence. In particular, we develop explanations for the LLaVa~\cite{llava} MLLM. While our methods are not strictly limited to this model alone, 
we choose LLaVa due to its performance and licensing.

\zeropar{Open-world Localization}\label{sec:background-owl}
Given an image $\mathbf{x}$ and a set of text queries $\mathbf{Q} = \{q_i\}$, open-world localization (OWL) consists in locating all instances of objects in an image that can be described by one of the queries in $\mathbf{Q}$. In practice, the output of an OWL model $\mathcal{D}$ is a set of bounding boxes $\mathcal{D}(\mathbf{x}) = \{(\mathbf{z},x,y,w,h)_i\}$, identifying \textit{i}) similarities $\mathbf{z}_i\in[0,1]^{|\mathbf{Q}|}$ over the queries to which the object may refer to, and \textit{ii}) its spatial location $x,y,w,h$ within the image.
In our work, we employ the OWL-ViT~\cite{owlvit-t} model, which achieves state-of-the-art performance in OWL benchmarks and competitive performance in long-tailed object detection.


\zeropar{Saliency Maps}\label{sec:background-saliency}
One of the most popular and established method to explain CV models is saliency maps. Given an image $\textbf{x}$, a saliency map is a heatmap $\textbf{y}$, with same size as the input image, that highlights which region of the image is most relevant for the model's prediction.
Saliency maps are typically computed with respect to a particular output, such as a particular output class $\mathbf{\omega}$. For example, the saliency map $\textbf{y}^{\mathbf{\omega}}$ explaining class $\mathbf{\omega}=\text{``\texttt{dog}''}$ will highlight the part of the image that most contributed to the output unit related to ``\texttt{dog}''.

For transformers, Attention Visualization~\cite{vit} and Attention Rollout~\cite{attentionrollout} are the prominent methodology enabling saliency map explanations. These, however, are limited to explaining the output of a ViT classifier for a particular output class, and are not suitable to explain an output token of a MLLM that is stacked on top of the ViT to be explained, as is the case in MLLM architectures. 
In our work, we exploit the proposed joint OWL-MLLM architecture to enable model explanations with respect to any output token at any position (Figures~\ref{fig:teaser},~\ref{fig:saliency}). 


\vspace{-7pt}
\section{Methods}
\vspace{-3pt}
We construct a joint Open-world Localization (OWL) and Multi-modal Large Language Model (MLLM) architecture $\mathcal{J}$ by combining and aligning OWL-ViT's \cite{owlvit-t} vision encoder with LLaVa's \cite{llava} language model.
This enables us to obtain a MLLM that serves both as language model and as an object detection model, displaying bounding boxes that enable us to visualize the model's understanding of the input image (Figure~\ref{fig:teaser}). In turn, this enables us to develop a saliency methodology to explain any output token (Section~\ref{sec:method-ga}), to detect hallucinations (Section~\ref{sec:exp-hallucination}), and to design adversarial perturbations to assess and measure model biases (Section~\ref{sec:method-bias}).

\zeropar{Combining the Models}\label{sec:method-combining}
We combine LLaVa~\cite{llava} and OWL-ViT~\cite{owlvit-t} and propose model $\mathcal{J}$ that outputs both text and bounding boxes given one multi-modal input (Figure~\ref{fig:teaser}). 
Crucially, in our design, we ensure that the vision representation is shared across language and detection outputs, such that \emph{perceived image semantics are identical for the two outputs}. Thus, the strong link between the two outputs can be exploited to interpret the model.

LLaVa~\cite{llava} employs a pre-trained CLIP~\cite{clip} vision encoder $\mathcal{E}^I$ and the LLaMa~\cite{llama} language model.
To predict new tokens, the encoding of the image $\mathcal{E}^{I}(\mathbf{x}) = \mathbf{t}_1^I,\dots,\mathbf{t}_{576}^I$ and of the prompt $\mathcal{T}(\mathbf{p}) = \textbf{t}_1,\dots,\textbf{t}_n$ are concatenated and sent to LLaMa.
OWL-ViT, instead, is composed of CLIP image and text encoders $\mathcal{E}^{OWL}$ that are trained from scratch, on top of which detection heads $\mathcal{H}$ provide an open-world localization output $\{(\mathbf{z},x,y,w,h)_i\}, i\in1,\dots,576$ (Section~\ref{sec:background-owl}), amounting to one box for each visual token. During normal operations, the $576$ boxes are filtered by thresholding on $\mathbf{z}_i$ to only display objects present in the image.

To construct $\mathcal{J}$, we replace LLaVa's vision encoder $\mathcal{E}^I$ with OWL-ViT's vision encoder $\mathcal{E}^{OWL}$, and train a new alignment MLP ${W}$ to take the output $\mathcal{E}^{OWL}(\mathbf{x}) = \mathbf{t}_1^{OWL},\dots,\mathbf{t}_{576}^{OWL}$ and transform it into a sequence ${W}(\mathcal{E}^{OWL}(\mathbf{x})) = \mathbf{t}_1^{{W}},\dots,\mathbf{t}_{576}^{{W}}$ that is compatible with the language model (Figure~\ref{fig:teaser}). LLaMa and OWL-ViT are otherwise kept frozen to retain localization and language modeling performance.
We train ${W}$ from scratch in a self-supervised manner over the entire Open Images~\cite{OpenImages} dataset, minimizing the loss:
\vspace{-5pt}
\begin{equation}
    \mathcal{L}({W}, \mathbf{x}) = |\mathcal{E}^{I}(\mathbf{x}) - {W}(\mathcal{E}^{OWL}(\mathbf{x}))|_{L2}.
    \label{eq:Wloss}
    \vspace{-5pt}
\end{equation}
We discuss training details in Section~\ref{sec:exp-W}.

\begin{algorithm*}[h]
    \begin{algorithmic}[1]
        \Require $\mathbf{x}$, $\mathbf{p}$, $\mathbf{Q}$
        \State $\textbf{t}^{OWL}_1,\dots,\textbf{t}^{OWL}_{576} = \mathcal{E}^{OWL}(\mathbf{x})$ \Comment{Embed image with OWL-ViT image encoder}
        \State $\textbf{t}^{{W}}_i = {W}(\textbf{t}^{OWL}_i), i\in 1,\dots,576$ \Comment{Align embedding using ${W}$}
        \State $\textbf{t}_1,\dots,\textbf{t}_n = \mathcal{T}(\mathbf{p})$ \Comment{Encode text with LLaVa's original text encoder}
        \State $\hat{\textbf{l}}_1,\dots,\hat{\textbf{l}}_{576+n} = \text{LLaMa}(\textbf{t}^{{W}}_1,\dots,\textbf{t}^{{W}}_{576},\textbf{t}_1,\dots,\textbf{t}_n)$ \Comment{Run LLaMa on the concat. encodings}
        \State $\{(\mathbf{z},x,y,w,h)_i\} = \mathcal{H}(\textbf{Q}, \textbf{t}^{OWL}_i), i\in1,\dots,576$ \Comment{Run object detection}
        \\ \Return $\textbf{O}^{MLLM} = \hat{\textbf{l}}_{576+n}$, $\textbf{O}^{OWL} = \{(\mathbf{z},x,y,w,h)_i\}$
    \end{algorithmic}
    \caption{$\mathcal{J}$ pipeline execution}
    \label{algo:model}
\end{algorithm*}

In Algorithm~\ref{algo:model}, we detail the procedure to run $\mathcal{J}$.
Given input image $\textbf{x}$, a list of text queries $\textbf{Q}$ and prompt $\textbf{p}$,
we first obtain the image's OWL-ViT embedding (\texttt{line 1}), and align it to the LLM via our proposed ${W}$ (\texttt{line 2}). The text is also encoded using LLaVa's original module $\mathcal{T}$ (\texttt{line 3}). The vision and text tokens are concatenated and fed to LLaMa (\texttt{line 4}) obtaining one logit vector per input token, the last of which ($\hat{\textbf{l}}_{576+n}$) is the one of interest for text generation ($\textbf{O}^{MLLM}$). We then run object detection with queries $\mathbf{Q}$, obtaining output $\textbf{O}^{OWL} = \{(\mathbf{z},x,y,w,h)_i\}$, and return $\textbf{O}^{MLLM}$, $\textbf{O}^{OWL}$ (\texttt{lines 5-6}). 
Thus, we have obtained language and bounding box outputs from the same vision representation where, notably, we can control the set of queries $\mathbf{Q}$ without altering the MLLM output. In turn, this enables to display the MLLM's perceived visual semantics of $\mathbf{x}$ with respect to any concept. 

\zeropar{Detecting Hallucinations}\label{sec:method-hallucination} An important consequence of the shared OWL and MLLM vision embeddings in $\mathcal{J}$ is that erroneous perceptions are reflected both in the text and in the detection outputs. Therefore, the proposed model enables the visualization of MLLM hallucinations via the detection output, greatly enhancing the interpretability of the model in case of errors. 
As shown in Figures~\ref{fig:teaser},~\ref{fig:hallucination}, when the MLLM outputs text referring to objects that are not present in the image, then the detection output also displays these objects.
In Section~\ref{sec:exp-hallucination} we design an experiment to confirm that this property holds.



\zeropar{Gradient Alignment Saliency Map}\label{sec:method-ga}
Given the proposed model $\mathcal{J}$, we develop the Gradient Alignment (GA) saliency map to explain the MLLM output. For input sequence $\textbf{t}^{{W}}_1,\dots,\textbf{t}^{{W}}_{576},\textbf{t}_1,\dots,\textbf{t}_n$, GA explains the last generated token $\hat{\textbf{t}}_{576+n} \doteq \textbf{t}_{576+n+1}$ with respect to a visual concept $c$ described via text (e.g., ``\texttt{dog}''). 

\begin{algorithm*}[h]
    \begin{algorithmic}[1]
        \Require $\mathbf{x}$, $\mathbf{p}$, $c$
        \State $\hat{\textbf{l}}, \{({z},x,y,w,h)_i\} = \mathcal{J}(\mathbf{x}, \mathbf{p}, \{c\})$ \Comment{Run model}
        \State $s = \argmax_r(\hat{\textbf{l}}[r])$ \Comment{Find top logit index}
        \State $\nabla_t = \frac{\partial \hat{\textbf{l}}[s]}{\partial \textbf{t}^{OWL}_1,\dots,\textbf{t}^{OWL}_{576}}$ \Comment{Gradient of top logit w.r.t. vision representation}
        \State $\nabla_{o, i} = \frac{\partial {z}_i}{\partial \textbf{t}^{OWL}_1,\dots,\textbf{t}^{OWL}_{576}}, i\in1,\dots,576$ \Comment{Gradient of $z_i$ w.r.t. vision representation}
        \State $r_i = \phi(\nabla_t, \nabla_{o, i})), i\in1,\dots,576$ \Comment{Compute box relevance scores}
        \State $\mathbf{y}^c \leftarrow \mathbf{-\infty}$ \Comment{Initialize output}
        \State \parbox{205pt}{$\mathbf{y}^c[y_i:y_i+h_i,x_i:x_i+w_i] =$ \\ $\hspace*{10pt} \text{max}(\mathbf{y}^c[y_i:y_i+h_i,x_i:x_i+w_i], r_i), i\in1,\dots,576$}
        \Comment{Compile output}
        \\ \Return $\text{normalize}(\mathbf{y}^c)$
    \end{algorithmic}
    \caption{Saliency map generation}
    \label{algo:saliency}
\end{algorithm*}

The main idea behind the proposed saliency map is to identify the relevance of each bounding box $\mathbf{o}_m$ to the output token $\hat{\mathbf{t}}_{576+n}$ by measuring the correlation between the gradients of the two outputs $\textbf{O}^{MLLM}, \textbf{O}^{OWL}$ with respect to the shared image embedding $\mathbf{t}_i^{OWL}$ (Algorithm \ref{algo:model}).
For example, if a particular LLM logit $\hat{\textbf{l}}_n$ is sensitive to a particular perturbation (e.g., the value for token ``\texttt{cat}'' decreases), then we also expect a particular output box $\mathbf{o}_m$ to be similarly sensitive (e.g., the value for query ``\texttt{cat}'' in $\mathbf{z}_m$ decreases).
Crucially, since both outputs are computed from the same vision representation, \emph{gradient correlation also implies semantic correlation}.
To assess this correlation, we measure the cosine similarity $\phi(\frac{\partial \mathbf{z}_m}{\partial\mathcal{E}^{OWL}}, \frac{\partial\hat{\textbf{l}}_n[s]}{\partial\mathcal{E}^{OWL}})$ where $\mathbf{z}_m$ indicates the relevance of the box to query $c$. If this value is high, then the image region covered by the $m$-th bounding box in $\textbf{O}^{OWL}$ is relevant to the output logit $\textbf{l}_n[s]$ in $\textbf{O}^{MLLM}$.
Most importantly, the box coordinates $x_i,y_i,w_i,h_i$ given by OWL are independent of the query, and thus \emph{the position of the boxes used for the saliency map are unbiased by the use of the concept $c$ as query}.

We now detail the procedure, following Algorithm~\ref{algo:saliency}.
First, we run $\mathcal{J}$ with a single query $\{c\}$ (thus $\mathbf{z}_i$ are scalars $z_i$),
and obtain the language and detection outputs (\texttt{line 1}). In particular, the output token is determined by the maximum value in the logit vector $\mathbf{\hat{l}}$, $\mathbf{\hat{l}}[s]$ (\texttt{line 2}). 
Then, we compute the gradients of the outputs of $\mathcal{J}$ with respect to the vision embedding, obtained as per Algorithm \ref{algo:model} (\texttt{lines 3-4}). Subsequently, for each output box, we compute the cosine similarity $\phi$ between the gradient $\nabla_t$ of the top language logit and $\nabla_{o,i}$ of the $i$-th box class logit for query $c$ with respect to the shared embedding $\mathbf{t}_i^{OWL}$, obtaining scores $r_i$ (\texttt{lines 5-6}). Lastly, the saliency map is obtained by assigning to each pixel the value of the maximum score $r_i$ for all boxes that cover that pixel (\texttt{lines 7-9}). The saliency map is thus normalized and returned (\texttt{line 10}). 
Notably, GA can explain tokens at any output position. To do so, it is sufficient to run the model until the desired token is the one being generated.
In Figure~\ref{fig:saliency}, we show examples of GA, computed with respect to each object in the image appearing in the output tokens. Additional examples and details are available in the supplementary material.

\zeropar{Bias assessment}\label{sec:method-bias}
Many social biases have been identified in LLMs~\cite{mllmbias},
a portion of which also affect MLLMs, such as \textit{biological gender} and \textit{ethnicity} biases, as these properties can be deduced from images.
For example, a MLLM prompted with ``\textit{From 1 to 10, how likely is this person to be a nurse?}'' may be biased to output higher scores when the input image depicts a female person, and lower scores when the input depicts a male person.


For the proposed $\mathcal{J}$, we can exploit the shared vision representation $\mathbf{t}_i^{OWL}$ to measure these effects. To do so, we design perturbations $\mathbf{e}$ applied to $\mathbf{t}_i^{OWL}$ to induce a particular effect on $\mathbf{O}^{OWL}$ and, by analyzing its effects on $\mathbf{O}^{MLLM}$, we can gain insight on the MLLM's functioning.
Given our ``\textit{nurse}'' example, for an image $\mathbf{x}$ depicting a \textit{female person}, we may design a perturbed vision representation $\mathbf{t}_i^{OWL} + \mathbf{e}$ such that $\mathbf{O}^{OWL}$ incorrectly identifies a \textit{male person}. Since the vision representation is shared, the alteration in the visual semantics will also be reflected in $\mathbf{O}^{MLLM}$. In particular, if the model is susceptible to biological gender bias, it follows that the model's answer score after the perturbation will be lower, since the model interprets a male to be less likely to be a nurse than a female.

\begin{algorithm*}[h]
    \begin{algorithmic}[1]
        \Require $\mathbf{x}$, $\mathbf{C}^+$, $\mathbf{C}^-$, $\delta$
        \State $\textbf{t}^{OWL}_1,\dots,\textbf{t}^{OWL}_{576} = \mathcal{E}^{OWL}(\mathbf{x})$ \Comment{OWL-ViT vision embedding}
        \State $\{(\mathbf{z},x,y,w,h)_i\} = \mathcal{H}(\textbf{t}^{OWL}_i, \mathbf{C}^- \cup \mathbf{C}^+), i\in1,\dots,576$ \Comment{Detection output}
        \State $\mathcal{L}_{ADV} = \sum_{i\in1,\dots,576}{\sum_{j\in1,\dots,|\mathbf{C}^-|}{\mathbf{z}_i[j]}} - \sum_{i\in1,\dots,576}{\sum_{j\in|\mathbf{C}^-|,\dots,|\mathbf{C}^-|+|\mathbf{C}^+|}{\mathbf{z}_i[j]}}$
        \State $\mathbf{e} = -\delta \cdot \text{sign}(\frac{\partial \mathcal{L}}{\partial \textbf{t}^{OWL}_1,\dots,\textbf{t}^{OWL}_{576}})$ \Comment{Compute FGSM perturbation}
        \\ \Return 
        $\textbf{t}^{ADV}_1,\dots,\textbf{t}^{ADV}_{576} = (\textbf{t}^{OWL}_1,\dots,\textbf{t}^{OWL}_{576}) + \mathbf{e}$ \Comment{Apply perturbation and return} 
    \end{algorithmic}
    \caption{Semantic adversarial attack}
    \label{algo:adversarial}
\end{algorithm*}

To compute $\mathbf{e}$, we design adversarial perturbations~\cite{intriguing} which, differently from previous works~\cite{adversarialsurvey}, can reflect a semantically meaningful change in the embedding, such that objects pertaining to a set of concepts $\mathbf{C}^-$ are substituted by concepts pertaining to a set $\mathbf{C}^+$ in the visual representation $\mathbf{t}_i^{OWL}$ (Figure~\ref{fig:teaser}).
Our method, described in Algorithm \ref{algo:adversarial}, is based on the Fast Gradient Sign Method (FGSM)~\cite{fsgm}, and is enabled by the compact  perceived image semantics representation in $\mathcal{J}$, which is described by very few parameters $\{(\mathbf{z}, x, y, w, h)_i\}$.

To perform the perturbation, we first compute the original OWL vision embedding (\texttt{line 1}) and detection output with queries in $\mathbf{C}^- \cup \mathbf{C}^+$ (\texttt{line 2}). Then, we compute the adversarial loss $\mathcal{L}_{ADV}$ (\texttt{line 3}) where a positive term takes into account all detection logits for concepts in $\mathbf{C}^-$, such that they are minimized, and a negative term considers detection logits for concepts in $\mathbf{C}^+$, such that they are maximized.
Then, we apply the FGSM~\cite{fsgm} attack (\texttt{line 4}) and compute perturbation $\mathbf{e}$ in the direction of loss minimization:
\vspace{-3pt}
\begin{equation}
    \mathbf{e} = -\delta \cdot \text{sign}(\frac{\partial \mathcal{L}}{\partial \textbf{t}^{OWL}_1,\dots,\textbf{t}^{OWL}_{576}}),
\vspace{-3pt}
\end{equation}
where $\delta > 0$ is a user-defined magnitude.
The perturbation is finally added to the original embedding, thus constituting the adversarial embedding which is returned (\texttt{line 5}).
Importantly, except for the desired change, $\mathbf{t}_i^{OWL}$'s semantics are unmodified, such that effects to $\mathbf{O}^{MLLM}$ can solely be attributed to the substitution of concepts $\mathbf{C}^-$, $\mathbf{C}^+$.
We evaluate $\mathcal{J}$'s proneness to biases in Section~\ref{sec:exp-bias}.


\vspace{-10pt}
\section{Experiments and Results}
\vspace{-3pt}
\label{sec:experiments}
We now discuss our training procedures to construct model $\mathcal{J}$ (Section~\ref{sec:exp-W}) and detail the experiments that were carried out to show that $\mathcal{J}$ can be used to identify hallucinations  (Section~\ref{sec:exp-hallucination}), to validate the Gradient Alignment (GA) saliency map (Section~\ref{sec:exp-ga}), and to assess and measure bias proneness  (Section~\ref{sec:exp-bias}).

\zeropar{Evaluating ${W}$}\label{sec:exp-W}
We train the alignment MLP 
${W}$ using the proposed loss (Eq.~\ref{eq:Wloss}) on the full Open Images V4~\cite{OpenImages} dataset ($\approx9M$ images) for $1$ epoch, rescaling all samples to OWL's input dimension $768 \times 768$.
More details are available in the supplementary material.

We evaluate performance following the GPT-as-a-judge paradigm~\cite{llava} and ask GPT-4 Vision~\cite{gpt4} to rate the captions provided by $\mathcal{J}$. A high score indicates that the evaluated MLLM is able of good visual reasoning. For $\mathcal{J}$, this also means that the output of the alignment layer ${W}(\mathcal{E}^{OWL}(\mathbf{x}))$ is a good substitute for the original LLaVa vision encoder.
As a baseline, we measure GPT-4 ratings, for the same images, of ground truth captions and captions provided by the original LLaVa model.
As shown in Table~\ref{tab:gptjudge}, the explainability/performance tradeoff is clear in the slightly lower performance of $\mathcal{J}$ with respect to both LLaVa and GT captions. Nonetheless, as shown in Figure~\ref{fig:saliency}, $\mathcal{J}$ is capable of good image understanding and reasoning. More example outputs available in the supplementary material.


\begin{table}[t]
\caption{Evaluation of alignment layer using gpt-as-a-judge~\cite{llava}. Reporting the average GPT-4 score for $100$ COCO captions provided by GT, LLaVa, and $\mathcal{J}$.}
\label{tab:gptjudge}
\centering
\resizebox{0.5\linewidth}{!}{
\begin{tabular}{|l|ccc|}
\hline
Model: & COCO GT & LLaVa & Ours \\ \hline
Avg. score: & $6.9$ & $\mathbf{7.0}$ & $6.1$ \\ \hline
\end{tabular}
}
\vspace{-10pt}
\end{table}

\begin{figure}[t]
    \centering
    \resizebox{0.9\linewidth}{!}{
        \includegraphics{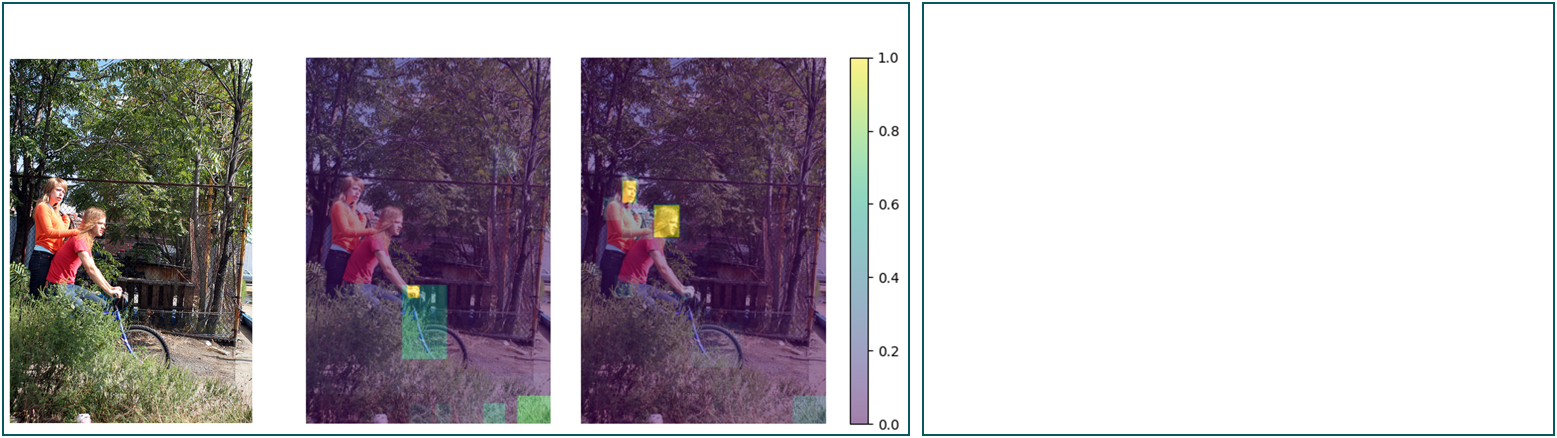}
    }
    \put(-327,84){\fontsize{8}{10}\selectfont Original image}
    \put(-261.5,84){\fontsize{8}{10}\selectfont Explanation 1}
    \put(-204,84){\fontsize{8}{10}\selectfont Explanation 2}
    \put(-130,70){\fontsize{8}{9}\selectfont \parbox{125pt}{Q: Which explanation is for object "bicycle" and which is for object "person"?}}
    \put(-130,37){\fontsize{8}{9}\selectfont \parbox{125pt}{A1: Explanation 1 explains "bicycle" and Explanation 2 explains "person"}}
    \put(-130,13){\fontsize{8}{9}\selectfont \parbox{125pt}{A2: Explanation 1 explains "person" and Explanation 2 explains "bicycle"}}
    
    \vspace{-5pt}
    \caption{Example question (Q) and answers (A1, A2) from the user study.}
    \vspace{-10pt}
    \label{fig:userstudy}
\end{figure}

\begin{table}[t]
\caption{Results for bias assessment.}
\label{tab:bias}
\centering
\resizebox{1.0\linewidth}{!}{
    \begin{tabular}{|l|lllc|}
    \hline
    \multicolumn{1}{|c|}{Bias} & \makecell[l]{\texttt{<property>} \\ \parbox{105pt}{\fontsize{8}{10}\selectfont $\rightarrow$ We ask the model to score from $1$ to $10$ the likelihood  of the person to possess this property}} & \makecell[l]{real characteristic ($\mathbf{C}^-$) \\ \parbox{100pt}{\fontsize{8}{10}\selectfont $\rightarrow$ The delta score is computed for the subset of images that possess this property \vspace{1pt}}} & \makecell[l]{adversarial characteristic ($\mathbf{C}^+$) \\ \parbox{125pt}{\fontsize{8}{10}\selectfont $\rightarrow$ The attack is carried out to reduce the likelihood of the real characteristic $\mathbf{C}^-$ in favour of the adversarial one $\mathbf{C}^+$ }} & \makecell{Delta score \\ \parbox{135pt}{\fontsize{8}{10}\selectfont $\rightarrow$ The difference in the model's numerical answer to the likelihood of \texttt{<property>} before and after the attack }} \vspace{1pt} \\ \hline
    \multirow{8}{*}{\begin{tabular}[c]{@{}l@{}}Biological gender \\ \parbox{185pt}{\fontsize{8}{10}\selectfont \,\\ $\rightarrow$ Stereotypical properties ``nurse'', ``construction worker'' are the MLLM answer to ``\textit{What is the most stereotypically} \texttt{<female|male>} \textit{job?}''. Delta scores for these properties highlight the model bias. \\ $\rightarrow$ Scores for ``female person'' and ``male person'' properties gauge the attack effectiveness.}\end{tabular}} & \multirow{2}{*}{nurse} & female person & male person & $-2.5$ \\
     &  & male person & female person & $+1.3$ \\ \cline{2-5} 
     & \multirow{2}{*}{construction worker} & female person & male person & $+0.3$ \\
     &  & male person & female person & $-1.4$ \\ \cline{2-5} 
     & \multirow{2}{*}{female person} & female person & male person & $-5.3$ \\
     &  & male person & female person & $+7.8$ \\ \cline{2-5} 
     & \multirow{2}{*}{male person} & female person & male person & $+6.8$ \\
     &  & male person & female person & $-2.0$ \\ \hline
    \multirow{8}{*}{\begin{tabular}[c]{@{}l@{}}Ethnicity \\ \parbox{185pt}{\fontsize{8}{10}\selectfont \,\\ $\rightarrow$ Stereotypical properties ``criminal'', ``rich'' are the MLLM answer to ``\textit{What stereotype for} \texttt{<african-american|caucasian>} \textit{people?}''. Delta scores for these properties highlight the model bias. \\ $\rightarrow$ Scores for ``african-american person'' and ``caucasian person'' properties gauge the attack effectiveness. }\end{tabular}} & \multirow{2}{*}{criminal} & african-american person & caucasian person & $-2.2$ \\
     &  & caucasian person & african-american person & $\pm0.0$ \\ \cline{2-5} 
     & \multirow{2}{*}{rich} & african-american person & caucasian person & \color{red} $-1.1$ \\
     &  & caucasian person & african-american person & \color{red} $\pm0.0$ \\ \cline{2-5} 
     & \multirow{2}{*}{african-american person} & african-american person &caucasian person & $-4.1$ \\
     &  & caucasian person & african-american person & $+0.9$ \\ \cline{2-5} 
     & \multirow{2}{*}{caucasian person} & african-american person & caucasian person & \color{red} $-2.2$ \\
     &  & caucasian person & african-american person & $-2.1$ \\ \hline
    \end{tabular}
} 
\vspace{-10pt}
\end{table}

\begin{table}[t]
\caption{Results for hallucination detection}
\label{tab:hallucinations}
\centering
\resizebox{0.8\linewidth}{!}{
\begin{tabular}{l|cc|}
\cline{2-3}
 & \multicolumn{2}{c|}{average $max_q$ for COCO images} \\
 & \multicolumn{1}{l}{MLLM answers ``\textit{Yes}''} & \multicolumn{1}{l|}{MLLM answers ``\textit{No}''} \\ \hline
\multicolumn{1}{|l|}{Response for GT classes} & $0.243$ & $0.113$ \\
\multicolumn{1}{|l|}{Response for negative classes} & $0.034$ & $0.017$ \\ \hline
\end{tabular}
}
\vspace{-10pt}
\end{table}

\zeropar{Hallucination Detection}\label{sec:exp-hallucination}
We design an experiment to verify that the strong link between $\mathbf{O}^{MLLM}$ and $\mathbf{O}^{OWL}$ can be exploited to detect whether the MLLM is hallucinating objects in the image (Section~\ref{sec:method-combining}).
Given a dataset composed of annotated images $\mathbf{x}_i$ containing objects of classes $\omega \in \Omega_i$, where $\Omega \supseteq \Omega_i$ is the label class set for the dataset, we ask the model ``\textit{Yes or no, does the image contain a} \texttt{<$\omega \in \Omega$>}\textit{?}'' and correlate the answers to the detection logits $q_\omega$ for query $\omega$.
We run the experiment on $1\,000$ COCO~\cite{coco} images and, for each class $\omega \in \Omega$, we record the model's \textit{Yes}/\textit{No} response and the maximal detection $max_q$:
\vspace{-6pt}
\begin{equation}
    max_q(\mathbf{x}, \omega) = max_i(\mathbf{z}_i), \{(\mathbf{z},x,y,w,h)_i\} = \mathcal{H}(\mathcal{E}^{OWL}(\mathbf{x}), \omega)
    \vspace{-6pt}
\end{equation}
For each image, we separate the computation between GT classes $\Omega_i$ and negative classes $\Omega \setminus \Omega_i$, as we expect two different distributions for $max_q$, higher for GT classes and lower for negative classes.

We report results in Table~\ref{tab:hallucinations}. It is clear that for classes where the model answers ``\textit{Yes}'', the average $max_q$ is higher, and lower when the model answers ``\textit{No}'', regardless of wether the object is present (GT class) or not (negative classes) in the image. This indicates a positive correlation between LLM ``\textit{Yes}'' answers and $max_q$, and thus the detection output.
Thus, we confirm that, when the MLLM hallucinates an object, the detection output will identify this non-present object with at least one high $\mathbf{z}_i$, as can be seen in the examples of Figure~\ref{fig:hallucination}. 

\zeropar{GA Saliency Map Evaluation}\label{sec:exp-ga}
A saliency map is only useful when it is understandable by its end users. As such, previous literature has introduced the concept of \textit{simulatability}~\cite{PV,PV1,PV8,PV12} to evaluate saliency methods, which is the property of saliency maps to enable the user to predict the model output. The idea is that, if the user is able to predict the output from the explanation of the input image, it follows that the saliency map is able to provide insight on the model's functioning. 

We assess simulatability through a user study, and ask users to predict which object in the image the explanation refers to (Figure~\ref{fig:userstudy}). Since GA explains the last output token, this amounts to effectively predicting the MLLM's output.
We setup the user study as an online questionnaire where users have to answer to $10$ questions. For each, they are presented with three versions of one PASCAL-VOC~\cite{voc2012} image, where the first is the original, and the others are overlay GA explanations of two ground truth objects in the image (Figure~\ref{fig:userstudy}).

We gather $17$ participants from Ms.C. and Ph.D. students with an AI background, and obtain average correct response rate of $0.941$, much higher than the expected random chance rate $0.5$ for $2$-option multiple choice questions. To validate this, we perform a one sample proportion binomial test where the null hypothesis is $H_0: p=0.5$. Given our sample size $17$ and proportion of correct answers $0.941$, we obtain p-value equal to $0.0001$, with test statistic $X=16$.
Thus, we reject $H_0$ and confirm that GA enables simulatability, and consequently we argue that our saliency methodology can help users gain insight on $\mathcal{J}$'s functioning.

\zeropar{Bias Assessment Benchmark}\label{sec:exp-bias}
Enabled by semantic adversarial perturbations (Section~\ref{sec:method-bias}), we propose a benchmark to measure the extent to which biases affect $\mathcal{J}$.
To do so, we construct datasets composed of $100$ portrait images for each of the considered biases, which are generated using SDXL~\cite{sdxl} to ensure that the images do not display objects other than the person. Furthermore, the datasets are split in two subsets, each displaying people pertaining to one of two categories object of the bias. For the biological gender bias, we generate $50$ images of women and $50$ images of men, for the ethnicity bias, we generate $50$ images of African-American people and $50$ images of Caucasian people.
Our selection of categories is not exhaustive, but was chosen to encompass diverse demographic representations relevant to the biases under study.
All samples are made publicly available, and generation details are discussed in the supplementary material.

As discussed in Section~\ref{sec:method-bias}, we can assess susceptibility to biases by running $\mathcal{J}$ twice for each image, once using embedding $\textbf{t}^{OWL}_1,\dots,\textbf{t}^{OWL}_{576} = \mathcal{E}^{OWL}(\mathbf{x})$, and once using the adversarial embedding $\textbf{t}^{ADV}_1,\dots,\textbf{t}^{ADV}_{576}$ (Algorithm~\ref{algo:adversarial}). For both embeddings, we use prompt ``\textit{From 1 to 10, how likely is this person to be} \texttt{<stereotypical property>}\textit{?}'', and measure changes in the answer before and after the attack.
Similar output changes could be discovered by running the model with images from the two subsets (e.g., obtaining lower scores for ``nurse'' with images of male people). However, 
the images could contain other contextual clues that may influence the MLLM's. We avoid these effects by analyzing score differences in relation to an attack constructed so that the only semantic alteration pertains to concepts in $\mathbf{C}^+, \mathbf{C}^-$, and by using SDXL-generated images that exclusively depict people's portraits.

For the two considered biases, we gather the \texttt{<stereotypical property>} by asking the model. This ensures that the considered characteristics reflect the model's reasoning and possible biases. For the biological gender bias, we ask the model ``\textit{What is the most stereotypically} \texttt{<female|male>} \textit{job?}'', and for the ethnicity bias, we ask ``\textit{What stereotype for} \texttt{<african-american|caucasian>} \textit{people?}''.

We display results, including the answers to these prompts, in Table~\ref{tab:bias}. 
When attacking with $\mathbf{C}^-={\text{``female person''}}$ and $\mathbf{C}^+={\text{``male person''}}$, scores for ``nurse'' decrease, while they increase in the reverse attack. For ``construction worker'', we see the opposite, with scores increasing when $\mathbf{C}^+={\text{``male person''}}$ and vice versa in the reverse attack. These results highlight the model's susceptibility to biological gender bias.
We also verify the successfulness of the attack by observing that scores for ``male person'' and ``female person'' properties changes accordingly when $\mathbf{C}^{+/-}=\{\text{"female person"}\},\{\text{"male person"}\}$.
For ethnicity, we discover that the attacks are not always successful, as demonstrated by the inconsistent results highlighted in red in Table~\ref{tab:bias}. In particular, scores for ``caucasian person'' decrease when $\mathbf{C}^-=\{\text{``african-american person"}\}$, $\mathbf{C}^+=\{\text{``caucasian person"}\}$. As such, results for stereotypical property ``rich'' are uninformative. Instead, for stereotypical property ``criminal'' the results follow the pattern described for the biological gender bias, demonstrating that the model is, to some extent, prone to ethicity bias.

\vspace{-7pt}
\section{Conclusions and Future Works}
\vspace{-3pt}
In conclusion, we have developed a novel architecture $\mathcal{J}$ by aligning the vision encoder of an OWL model to a MLLM. This enables to obtain a compact representation of the vision input, which in turn enables interpretability of the model via our proposed GA saliency map, hallucination visualization, and bias assessment through semantic adversarial perturbations.

Future works may improve the architecture by fine-tuning the entire MLLM on OWL vision encoding, thus overcoming the limitations of the alignment layer. Moreover, the GA explanation could be improved to also analyze the relationship between box position/size and text output, potentially unveiling interesting patterns. Lastly, our work could be extended to other forms of bias and to other MLLM architectures.







\bibliography{egbib}
\end{document}


\maketitle
\vspace{-6pt}

\section{Training $W$}
As discussed in Section 3.1, we train an alignment layer $\hat{W}$ between OWL's vision embedding and LLaVa's language model. The chosen architecture is a $2$-hidden layer MLP with hidden dimensionality $8192$. The input and output dimensionalities are $768$ and $4096$, to conform to OWL-ViT and LLaVa's token dimensionalities, respectively.
The alignment is applied to each vision token independently, such that given an input $B\times S\times 768$, we obtain output $B\times S\times 4096$, where the sequence length $S$ is typically $576$, which is the number of output vision tokens from OWL-ViT and equal to the sequence length required by the pre-trained LLaVa model.
In our work, we were limited in computational resources, but we hypothesize that by fine-tuning or training the LLaVa model without an alignment layer could be possible. In this case, the language model would learn to reason from the vision tokens as they are outputted from the OWL model. This would potentially lead to better performance, since the alignment inherently leads to information loss, as shown in the main paper, Table 1.
Additionally, we experimented with alternative architectures for the alignment layers, including the original LLaVa's alignment layer architecture and with a more complex transformer applied to the entire sequence, but with poor results, as shown in Table 1. Model weights are publicly available on~\href{https://github.com/loris2222/ExplainingMLLMs}{github}.

\zeropar{Evaluation details}
As discussed in the paper in Section 3.1, we follow the gpt-as-a-judge paradigm to evaluate alignment layers. We first generate captions from all models to be evaluted for the same $100$ random COCO images, using prompt:
\vspace{-1pt}
\begin{quote}
    \texttt{{\color{blue} USER:}} ``\textit{What's the content of the image?}''
\vspace{-0pt}
\end{quote}
We then prompt a strong judge (OpenAI's GPT4-vision) to score the provided captions:
\vspace{-1pt}
\begin{quote}
    \texttt{{\color{blue} SYSTEM:}} ``\textit{Please act as an impartial judge and evaluate the quality of the caption provided by an AI assistant for the provided image. Your evaluation should consider factors such as the helpfulness, relevance, accuracy, depth, and level of detail of the caption. Be as objective as possible.}''\\
    \texttt{{\color{blue} USER:}} ``\textit{From 1 to 10, score this caption for the image: <caption>. Only answer with a number from 1 to 10.}''\\
    \vspace{-1pt}
\end{quote}
We then average the $100$ GPT4-provided scores for the captions generated by each model to be evaluated, and obtain the results of Table 1.

\vspace{5pt}
\begin{table}[h]
\caption{Evaluation of alignment layer using gpt-as-a-judge. Reporting the average GPT-4 score for $100$ COCO captions provided by different models.}
\label{tab:supp-judge}
\centering
\resizebox{0.9\linewidth}{!}{
\begin{tabular}{|l|ccccc|}
\hline
Model: & COCO GT & LLaVa & Ours (deep MLP) & Ours (LLaVa MLP) & Ours (transformer) \\ \hline
Avg. score: & $6.9$ & $\mathbf{7.0}$ & $6.1$ & $5.6$ & $3.2$ \\ \hline
\end{tabular}
}
\end{table}

\section{Model output examples}
In the following figures, we present examples of model outputs to multi-modal inputs ($\mathbf{x}$, $\mathbf{p}$), demonstrating that the alignment layer enables the LLM to understand vision information provided by OWL-ViT.
Prompts are taken from LLaVA-Instruct-150K.
We highlight in red discrepancies between GT and original LLaVa output with $\mathcal{J}$ due to imperfect alignments. If there are any, we highlight the detection for the relevant objects. From Figures 1-4, it is clear how the detection output can help explain the LLM output.

\begin{figure}[htbp]
    \centering
    \begin{tabular}{c|c|c}
        \multicolumn{3}{c}{\includegraphics[width=0.7\textwidth]{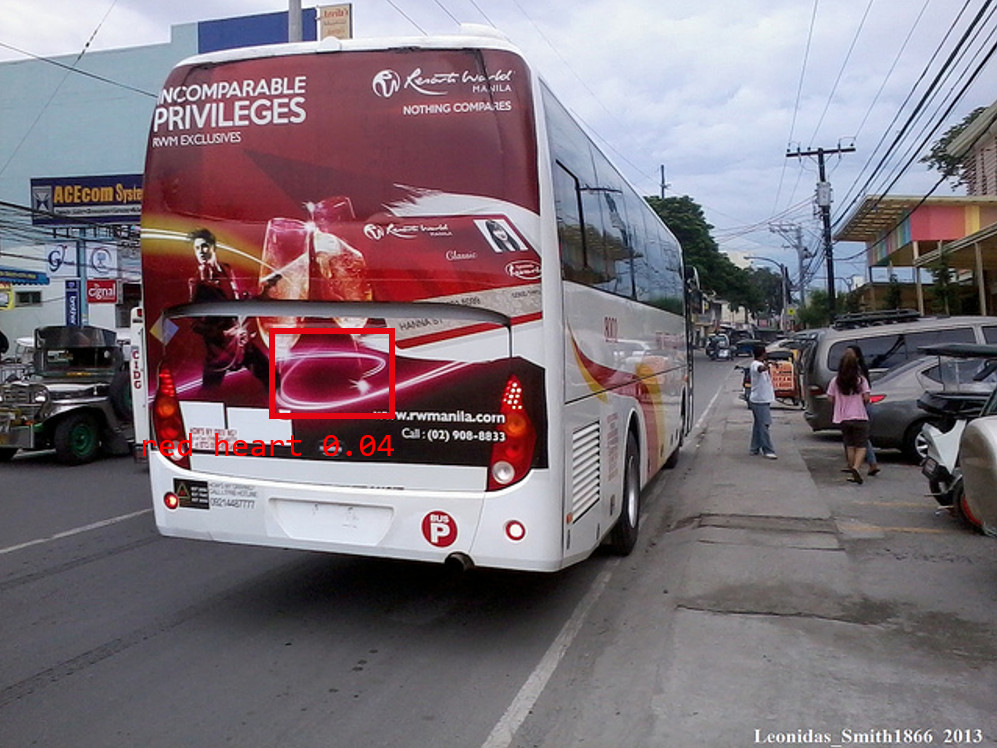}} \\
        \parbox{.3\linewidth}{GT: \\} &
        \parbox{.3\linewidth}{Original LLaVa: \\} &
        \parbox{.3\linewidth}{$\mathcal{J}$: \\} \\
        \parbox[t]{.3\linewidth}{\fontsize{8}{9}\selectfont USER: What are the colors of the bus in the image?\\ 
ASSISTANT: The bus in the image is white and red.\\ 
USER: What feature can be seen on the back of the bus?\\ 
ASSISTANT: The back of the bus features an advertisement.\\ 
USER: Is the bus driving down the street or pulled off to the side?\\ 
ASSISTANT: The bus is driving down the street, which is crowded with people and other vehicles.} &
        \parbox[t]{.3\linewidth}{\fontsize{8}{9}\selectfont USER: What are the colors of the bus in the image?\\ 
ASSISTANT: The bus in the image is red, white, and yellow.\\ 
USER: What feature can be seen on the back of the bus?\\ 
ASSISTANT: There is a large advertisement on the back of the bus.\\ 
USER: Is the bus driving down the street or pulled off to the side?\\ 
ASSISTANT: The bus is driving down the street.} &
        \parbox[t]{.3\linewidth}{\fontsize{8}{9}\selectfont USER: What are the colors of the bus in the image?\\ 
ASSISTANT: The bus in the image is red and white.\\ 
USER: What feature can be seen on the back of the bus?\\ 
ASSISTANT: The back of the bus has a {\color{red} large red heart}.\\ 
USER: Is the bus driving down the street or pulled off to the side?\\ 
ASSISTANT: The bus is driving down the street.} \\
    \end{tabular}
    \caption{Model $\mathcal{J}$ answers regarding a red heart instead of the advertisement on the back of the bus. As the detection output indicates, the graphic on the advertisement may resemble a red heart, which is identified by both $\mathbf{O}^{MLLM}$ and $\mathbf{O}^{OWL}$.}
    \label{fig:first-conv}
\end{figure}

\begin{figure}[htbp]
    \centering
    \begin{tabular}{c|c|c}
        \multicolumn{3}{c}{\includegraphics[width=0.8\textwidth]{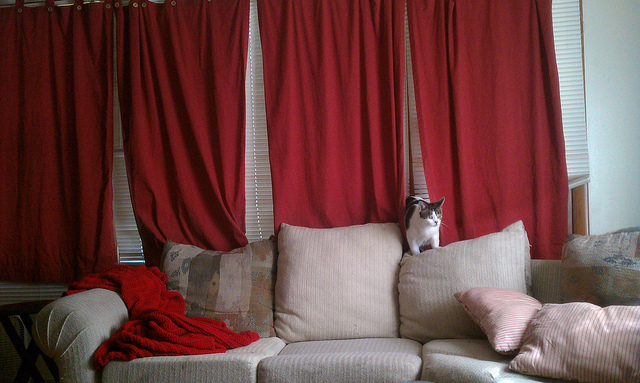}} \\
        \parbox{.3\linewidth}{GT: \\} &
        \parbox{.3\linewidth}{Original LLaVa: \\} &
        \parbox{.3\linewidth}{$\mathcal{J}$: \\} \\
        \parbox[t]{.3\linewidth}{\fontsize{8}{9}\selectfont USER: Where is the cat positioned in the image?\\ 
ASSISTANT: The cat is positioned on top of the back of the couch in the living room.\\ 
USER: What is the cat doing in the image?\\ 
ASSISTANT: The cat is coming out from some curtains onto the couch and is sitting or standing on top of it.\\ 
USER: What is the color of the curtains near the cat?\\ 
ASSISTANT: The color of the curtains near the cat is red.\\ 
USER: Besides the cat, what else can be seen on the couch?\\ 
ASSISTANT: There are lots of pillows on the couch besides the cat.\\ 
USER: What kind of room does the couch appear to be in?\\ 
ASSISTANT: The couch appears to be in a living room setting.} &
        \parbox[t]{.3\linewidth}{\fontsize{8}{9}\selectfont USER: Where is the cat positioned in the image?\\ 
ASSISTANT: The cat is positioned on top of a couch in the image.\\ 
USER: What is the cat doing in the image?\\ 
ASSISTANT: The cat is sitting or standing on top of the couch, possibly looking out the window.\\ 
USER: What is the color of the curtains near the cat?\\ 
ASSISTANT: The curtains near the cat are red.\\ 
USER: Besides the cat, what else can be seen on the couch?\\ 
ASSISTANT: There are pillows on the couch, in addition to the cat.\\ 
USER: What kind of room does the couch appear to be in?\\ 
ASSISTANT: The couch appears to be in a living room.} &
        \parbox[t]{.3\linewidth}{\fontsize{8}{9}\selectfont USER: Where is the cat positioned in the image?\\ 
ASSISTANT: The cat is positioned on a couch in the image.\\ 
USER: What is the cat doing in the image?\\ 
ASSISTANT: The cat is sitting or laying on the couch.\\ 
USER: What is the color of the curtains near the cat?\\ 
ASSISTANT: The color of the curtains near the cat is red.\\ 
USER: Besides the cat, what else can be seen on the couch?\\ 
ASSISTANT: A {\color{red} blanket} can also be seen on the couch along with the cat.\\ 
USER: What kind of room does the couch appear to be in?\\ 
ASSISTANT: The couch appears to be in a living room.} \\
    \end{tabular}
    \caption{Model $\mathcal{J}$ answers in almost an identical fashion to both GT and LLaVa. However, instead of mentioning the pillows on the couch, the model mentions a blanket, which is also present and an equally valid response.}
\end{figure}

\begin{figure}[htbp]
    \centering
    \begin{tabular}{c|c|c}
        \multicolumn{3}{c}{\includegraphics[width=0.8\textwidth]{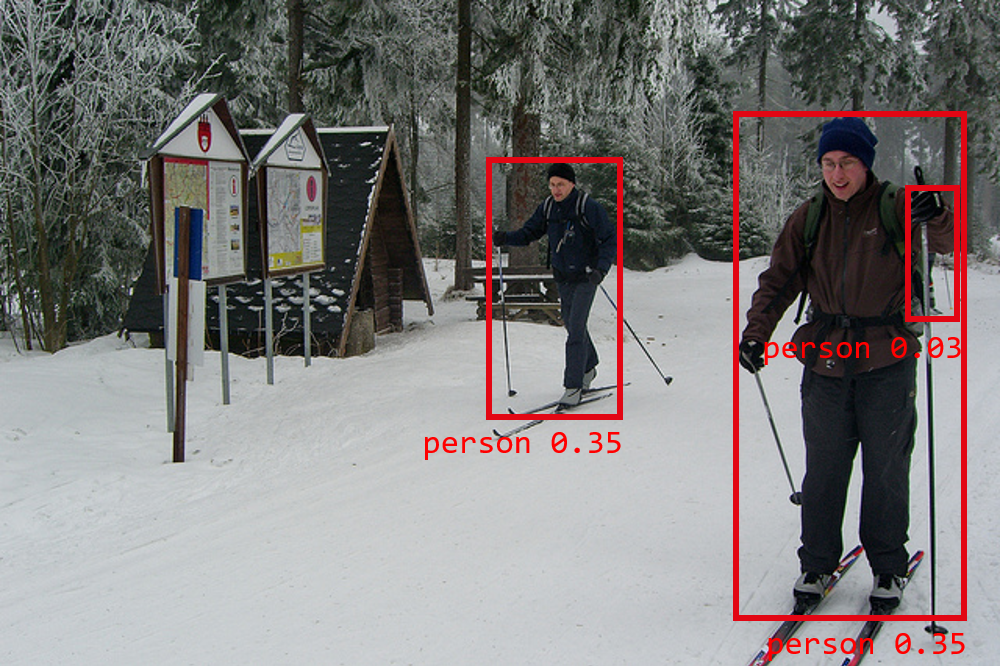}} \\
        \parbox{.3\linewidth}{GT: \\} &
        \parbox{.3\linewidth}{Original LLaVa: \\} &
        \parbox{.3\linewidth}{$\mathcal{J}$: \\} \\
        \parbox[t]{.3\linewidth}{\fontsize{8}{9}\selectfont USER: Are the people in the image skiing downhill or cross-country skiing?\\ 
ASSISTANT: The people in the image are cross-country skiing in the woods, as they are skiing on a trail rather than a steep slope.\\ 
USER: How many people are in the image?\\ 
ASSISTANT: There are two people in the image, both on skis in the snow.\\ 
USER: What kind of environment are they skiing in?\\ 
ASSISTANT: They are skiing in a wooded environment, following a trail through the trees while surrounded by snow.\\ 
USER: Do the skiers have any additional gear with them besides their skis and poles?\\ 
ASSISTANT: Yes, the two male skiers are carrying backpacks while they ski through the woods. The backpacks might contain essentials for their skiing adventure, such as food, water, extra clothing, or safety equipment.} &
        \parbox[t]{.3\linewidth}{\fontsize{8}{9}\selectfont USER: Are the people in the image skiing downhill or cross-country skiing?\\ 
ASSISTANT: The people in the image are cross-country skiing, as they are using skis and poles to move across the snow-covered ground.\\ 
USER: How many people are in the image?\\ 
ASSISTANT: There are two people in the image.\\ 
USER: What kind of environment are they skiing in?\\ 
ASSISTANT: The two people are skiing in a snowy environment, likely in a forest or a similar natural setting.\\ 
USER: Do the skiers have any additional gear with them besides their skis and poles?\\ 
ASSISTANT: Yes, the skiers have backpacks with them, which may contain essential items for their cross-country skiing adventure.} &
        \parbox[t]{.3\linewidth}{\fontsize{8}{9}\selectfont USER: Are the people in the image skiing downhill or cross-country skiing?\\ 
ASSISTANT: The people in the image are cross-country skiing, as they are standing on a snow-covered slope and holding their skis.\\ 
USER: How many people are in the image?\\ 
ASSISTANT: There are {\color{red} three} people in the image.\\ 
USER: What kind of environment are they skiing in?\\ 
ASSISTANT: They are skiing in a snow-covered environment, which could be a mountain or a slope.\\ 
USER: Do the skiers have any additional gear with them besides their skis and poles?\\ 
ASSISTANT: Yes, the skiers have backpacks with them, which could be used to carry additional supplies or equipment for their cross-country skiing adventure.} \\
    \end{tabular}
    \caption{Model $\mathcal{J}$ identifies three people in the image instead of two, as specified by the GT and in the LLaVa answer. However, as also indicated by $\mathbf{O}^{OWL}$, there is in fact a third person hidden behind the one on the right in foreground, as demonstrated by the barely visible ski and ski stick. We hypothesize that, since OWL-ViT is a more powerful detection model than the original LLaVa vision encoder, this information may have been lost with the original encoder. Instead, $\mathcal{J}$ was able to reason and answer also regarding this third person.}
\end{figure}

\begin{figure}[htbp]
    \centering
    \begin{tabular}{c|c|c}
        \multicolumn{3}{c}{\includegraphics[width=0.5\textwidth]{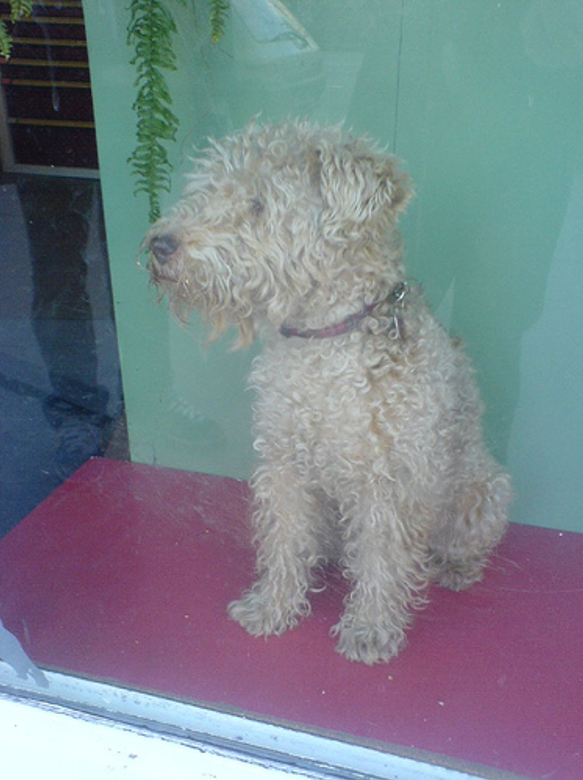}} \\
        \parbox{.3\linewidth}{GT: \\} &
        \parbox{.3\linewidth}{Original LLaVa: \\} &
        \parbox{.3\linewidth}{$\mathcal{J}$: \\} \\
        \parbox[t]{.3\linewidth}{\fontsize{8}{9}\selectfont USER: What color is the dog in the image?\\ 
ASSISTANT: The dog in the image is brown.\\ 
USER: What type of collar does the dog have?\\ 
ASSISTANT: The dog has a red collar.\\ 
USER: What is the dog's position behind the window?\\ 
ASSISTANT: The dog is sitting behind the window.\\ 
USER: How does the dog appear to feel as it looks out the window?\\ 
ASSISTANT: As the dog stares out the window, it appears to feel longingly, perhaps suggesting a desire to be outside or interacting with someone or something that has caught its attention.\\ 
USER: Can you describe the dog's fur?\\ 
ASSISTANT: The dog has shaggy, fuzzy, and furry brown fur, which makes it look cute and cuddly.} &
        \parbox[t]{.3\linewidth}{\fontsize{8}{9}\selectfont USER: What color is the dog in the image?\\ 
ASSISTANT: The dog in the image is brown.\\ 
USER: What type of collar does the dog have?\\ 
ASSISTANT: The dog has a leather collar.\\ 
USER: What is the dog's position behind the window?\\ 
ASSISTANT: The dog is sitting behind the window.\\ 
USER: How does the dog appear to feel as it looks out the window?\\ 
ASSISTANT: The dog appears to be feeling sad as it looks out the window.\\ 
USER: Can you describe the dog's fur?\\ 
ASSISTANT: The dog has long, shaggy fur.} &
        \parbox[t]{.3\linewidth}{\fontsize{8}{9}\selectfont USER: What color is the dog in the image?\\ 
ASSISTANT: The dog in the image is {\color{red}white}.\\ 
USER: What type of collar does the dog have?\\ 
ASSISTANT: The dog has a {\color{red} purple} collar.\\ 
USER: What is the dog's position behind the window?\\ 
ASSISTANT: The dog is sitting behind the window.\\ 
USER: How does the dog appear to feel as it looks out the window?\\ 
ASSISTANT: The dog appears to be feeling sad as it looks out the window.\\ 
USER: Can you describe the dog's fur?\\ 
ASSISTANT: The dog has {\color{red} fluffy, white fur}.} \\
    \end{tabular}
    \caption{Model $\mathcal{J}$'s output differs from the GT and LLaVa in terms of colors of the dog and collar and in the qualities of the dog's fur. Nonetheless, these are reasonable responses. Indeed, the collar is purplish-red, the dog is a very light brown color, and its fur is indeed fluffy.}
    \label{fig:last-conv}
\end{figure}

\FloatBarrier

\section{User Study Questions}
We now present all the $10$ questions that were asked to participants of the user study and report the number of correct responses for each question. These images also serve as additional examples of GA saliency maps.

\begin{longtable}{p{0.5\textwidth}p{0.4\textwidth}}
    \parbox[t]{0.4\textwidth}{\includegraphics[width=0.5\textwidth]{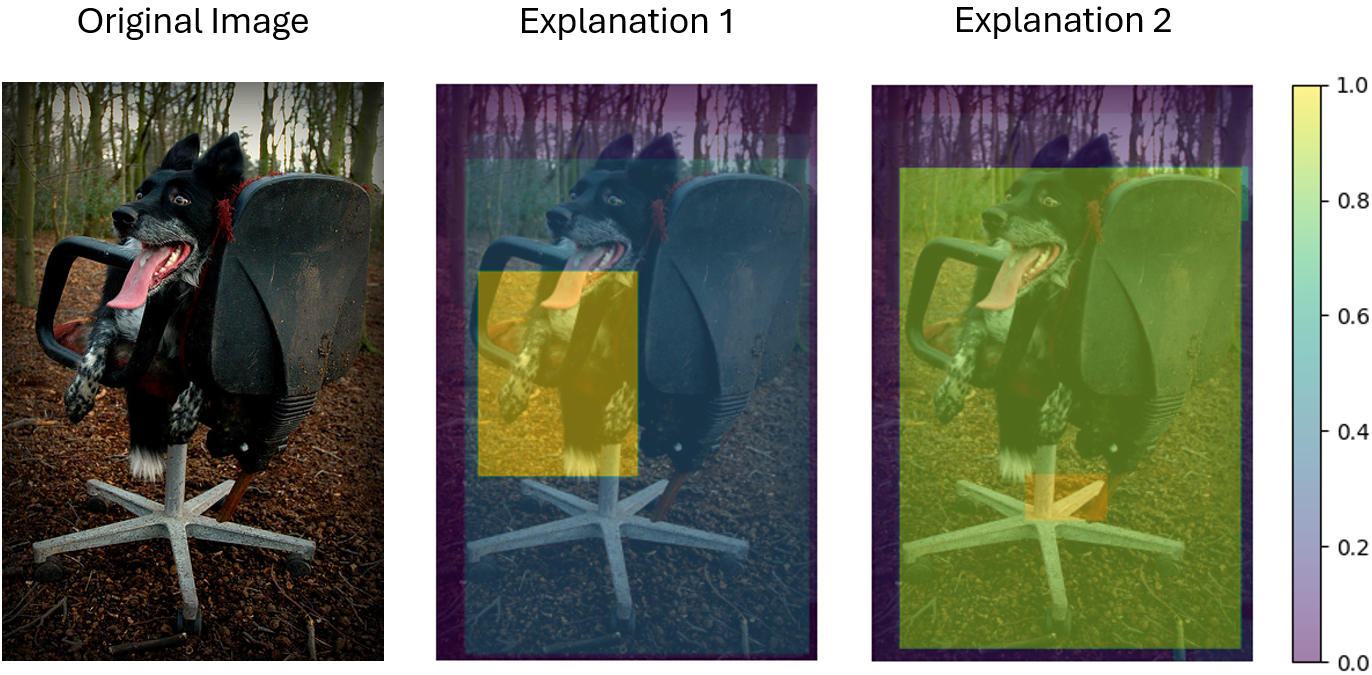}} & \parbox[t]{0.4\textwidth}{\vspace{-80pt}{\fontsize{7}{8}\selectfont Q: Which explanation is for object "chair" and which is for object "dog"? \\ A1: Explanation 1 explains "chair" and Explanation 2 explains "dog" \\ A2: Explanation 1 explains "dog" and Explanation 2 explains "chair" \\ Correct answer is A2 with $100\%$ correct answers.}}\vspace{10pt}\\
    \parbox[t]{0.4\textwidth}{\includegraphics[width=0.5\textwidth]{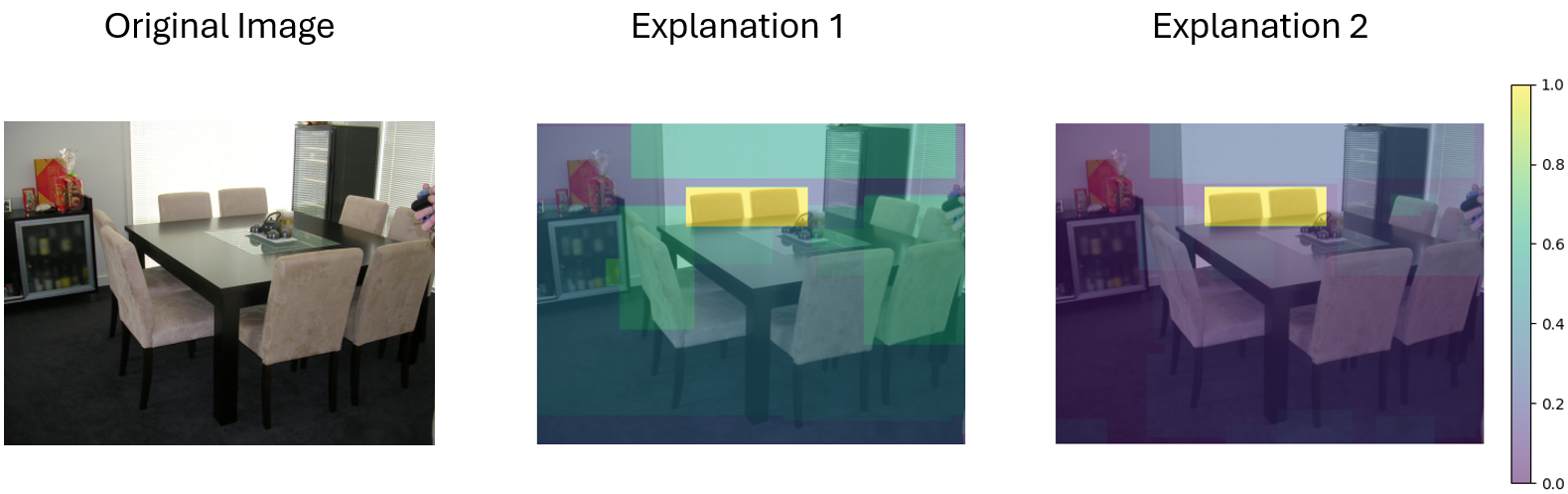}} & \parbox[t]{0.4\textwidth}{\vspace{-47pt}{\fontsize{7}{8}\selectfont Q: Which explanation is for object "chair" and which is for object "table"? \\ A1: Explanation 1 explains "chair" and Explanation 2 explains "table" \\ A2: Explanation 1 explains "table" and Explanation 2 explains "chair" \\ Correct answer is A2 with $71\%$ correct answers.}}\vspace{10pt}\\
    \parbox[t]{0.4\textwidth}{\includegraphics[width=0.5\textwidth]{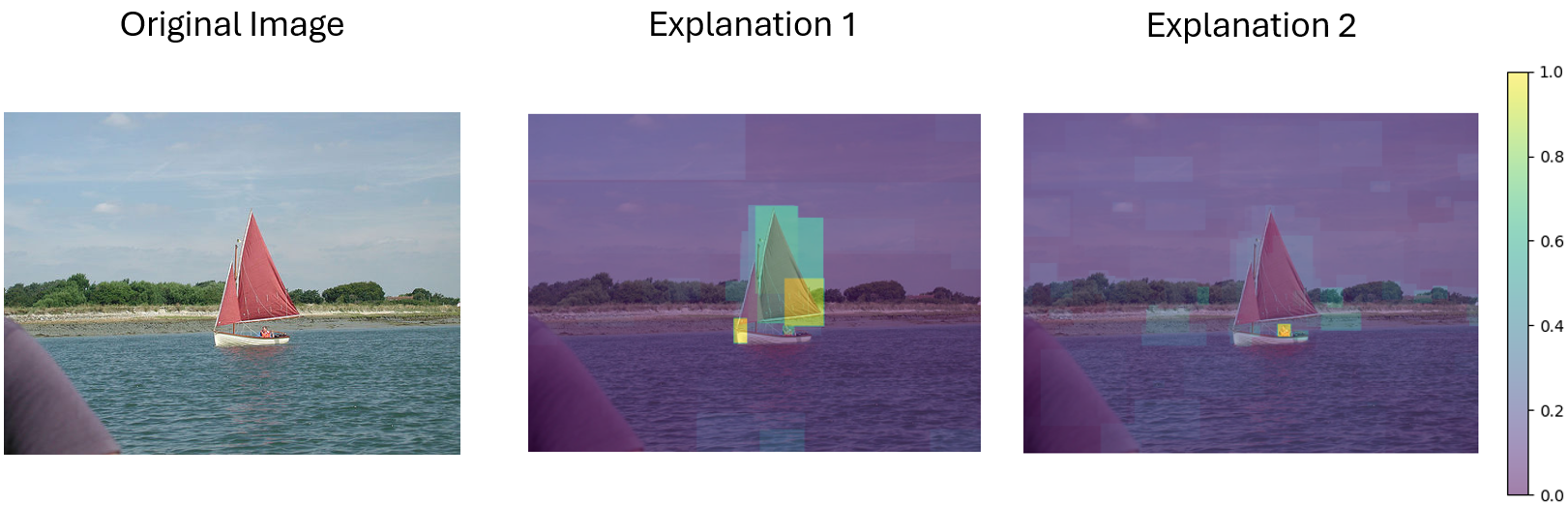}} & \parbox[t]{0.4\textwidth}{\vspace{-50pt}{\fontsize{7}{8}\selectfont Q: Which explanation is for object "boat" and which is for object "person"? \\ A1: Explanation 1 explains "boat" and Explanation 2 explains "person"	\\ A2: Explanation 1 explains "person" and Explanation 2 explains "boat" \\ Correct answer is A1 with $100\%$ correct answers.}}\vspace{10pt}\\
    \parbox[t]{0.4\textwidth}{\includegraphics[width=0.5\textwidth]{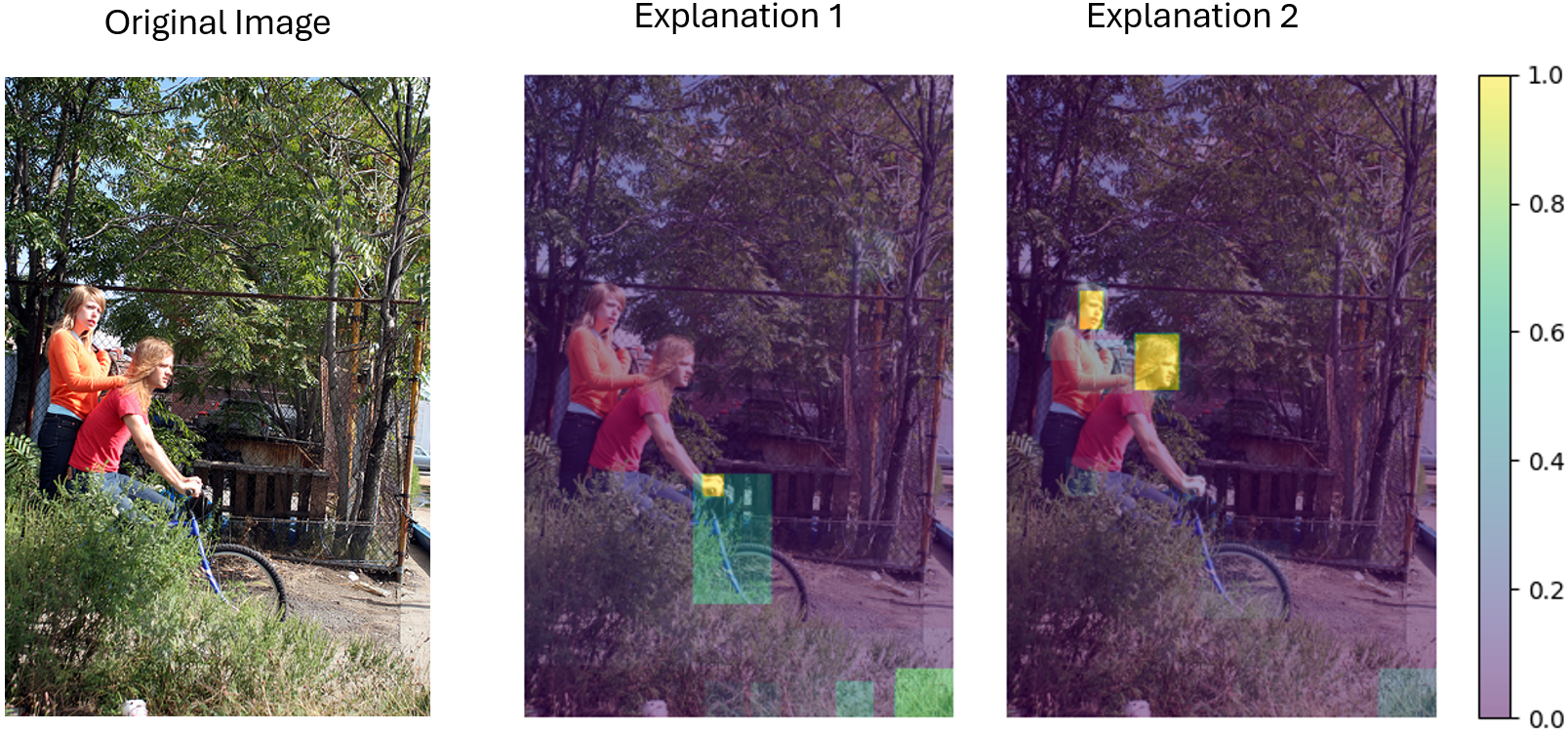}} & \parbox[t]{0.4\textwidth}{\vspace{-77pt}{\fontsize{7}{8}\selectfont Q: Which explanation is for object "bicycle" and which is for object "person"? \\ A1: Explanation 1 explains "bicycle" and Explanation 2 explains "person" \\ A2: Explanation 1 explains "person" and Explanation 2 explains "bicycle" \\ Correct answer is A1 with $100\%$ correct answers.}}\vspace{10pt}\\
    \parbox[t]{0.4\textwidth}{\includegraphics[width=0.5\textwidth]{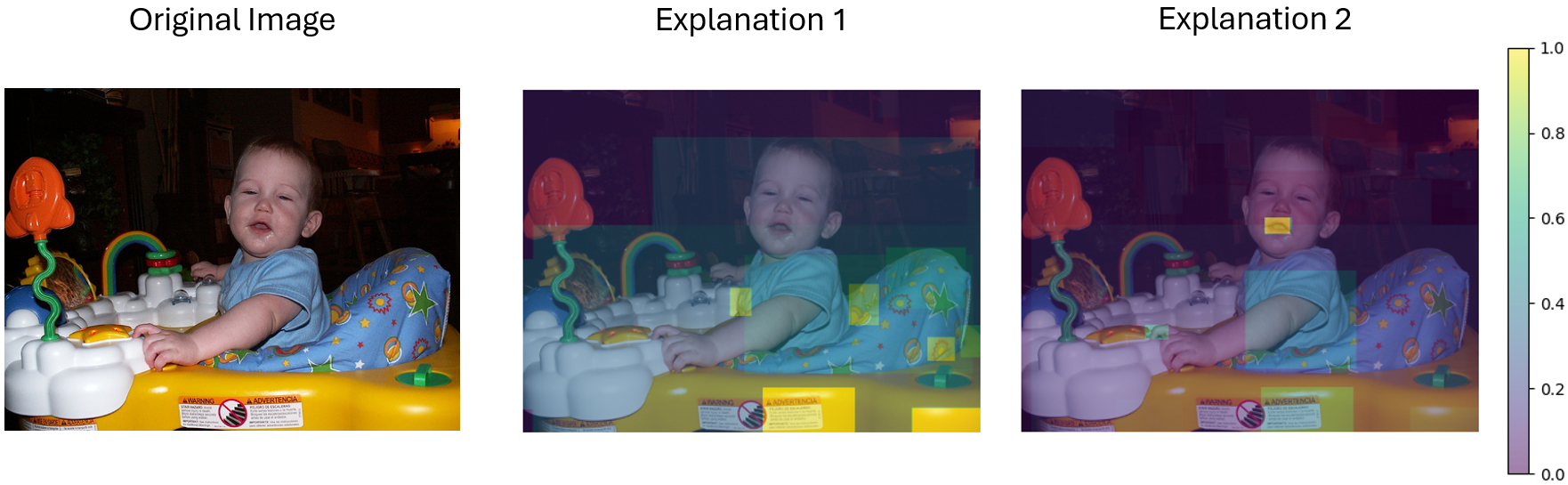}} & \parbox[t]{0.4\textwidth}{\vspace{-50pt}{\fontsize{7}{8}\selectfont Q: Which explanation is for object "person" and which is for object "chair"? \\ A1: Explanation 1 explains "person" and Explanation 2 explains "chair" \\ A2: Explanation 1 explains "chair" and Explanation 2 explains "person" \\ Correct answer is A2 with $100\%$ correct answers.}}\vspace{10pt}\\
    \parbox[t]{0.4\textwidth}{\includegraphics[width=0.5\textwidth]{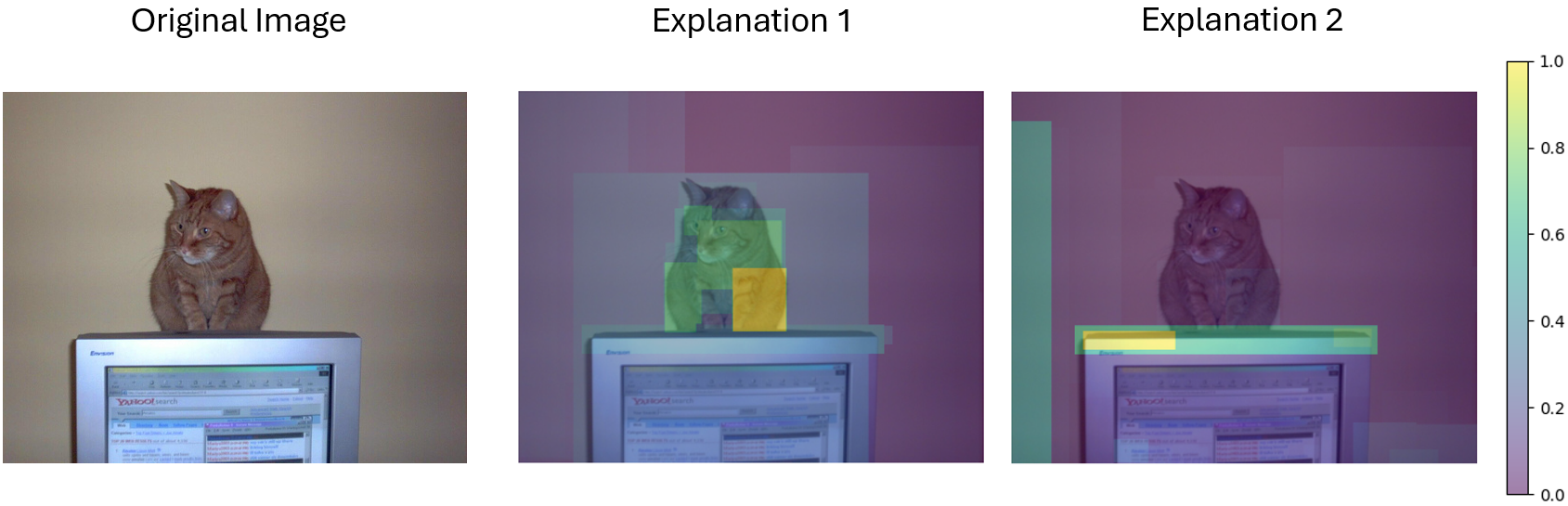}} & \parbox[t]{0.4\textwidth}{\vspace{-52pt}{\fontsize{7}{8}\selectfont Q: Which explanation is for object "cat" and which is for object "TV/monitor"? \\ A1: Explanation 1 explains "cat" and Explanation 2 explains "TV/monitor" \\ A2: Explanation 1 explains "TV/monitor" and Explanation 2 explains "cat" \\ Correct answer is A1 with $100\%$ correct answers.}}\vspace{10pt}\\
    \parbox[t]{0.4\textwidth}{\includegraphics[width=0.5\textwidth]{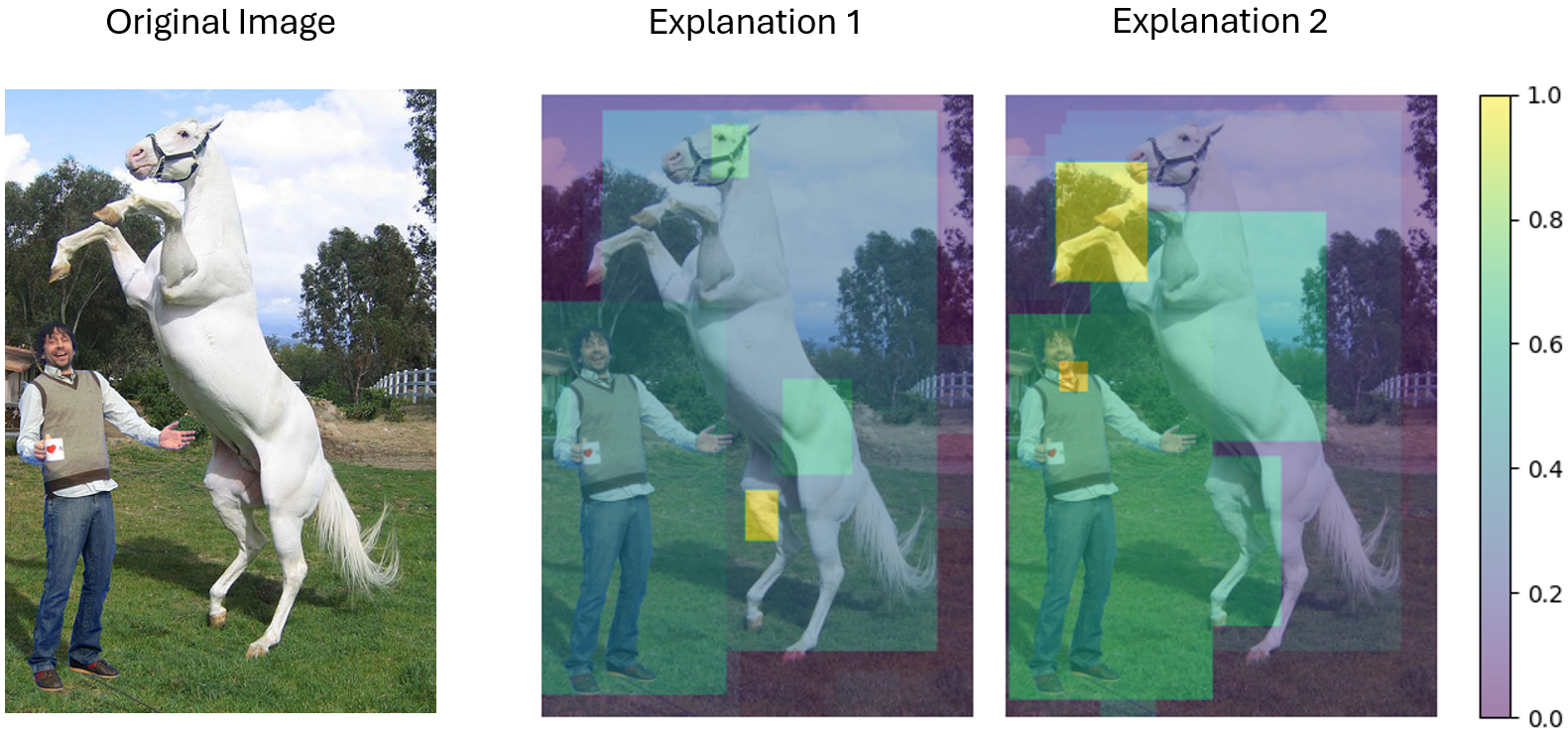}} & \parbox[t]{0.4\textwidth}{\vspace{-75pt}{\fontsize{7}{8}\selectfont Q: Which explanation is for object "person" and which is for object "horse"? \\ A1: Explanation 1 explains "person" and Explanation 2 explains "horse" \\ A2: Explanation 1 explains "horse" and Explanation 2 explains "person" \\ Correct answer is A2 with $71\%$ correct answers.}}\vspace{10pt}\\
    \parbox[t]{0.4\textwidth}{\includegraphics[width=0.5\textwidth]{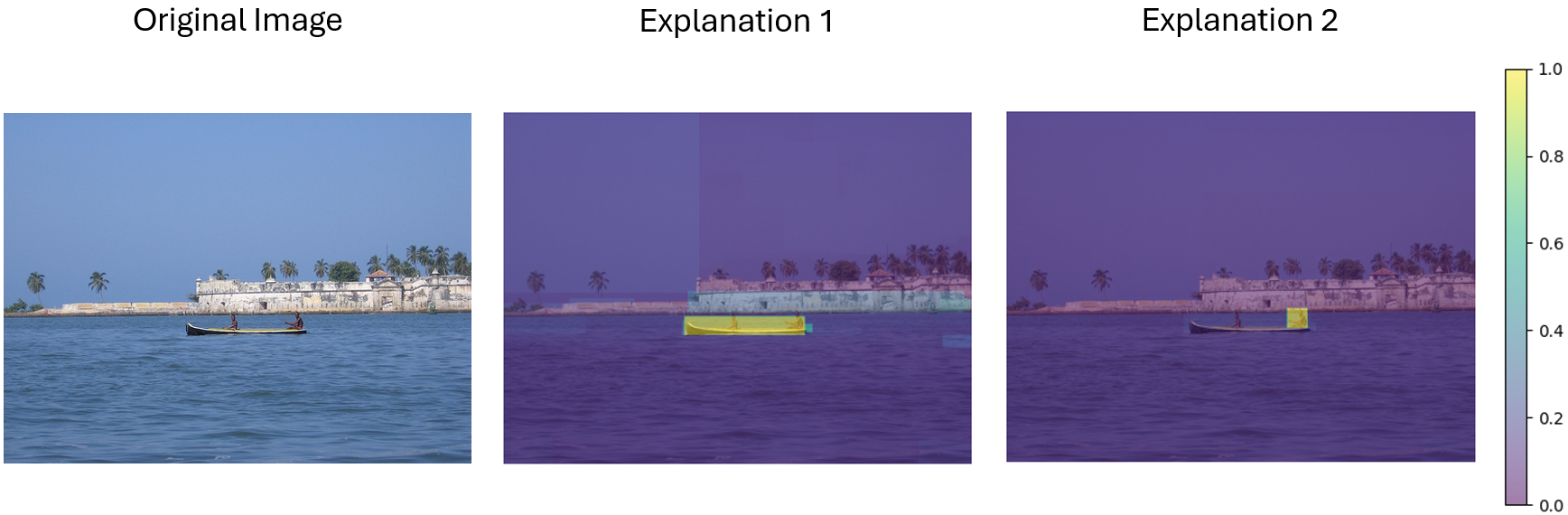}} & \parbox[t]{0.4\textwidth}{\vspace{-52pt}{\fontsize{7}{8}\selectfont Q: Which explanation is for object "person" and which is for object "boat"? \\ A1: Explanation 1 explains "person" and Explanation 2 explains "boat" \\ A2: Explanation 1 explains "boat" and Explanation 2 explains "person"\\ Correct answer is A2 with $100\%$ correct answers.}}\vspace{10pt}\\
    \parbox[t]{0.4\textwidth}{\includegraphics[width=0.5\textwidth]{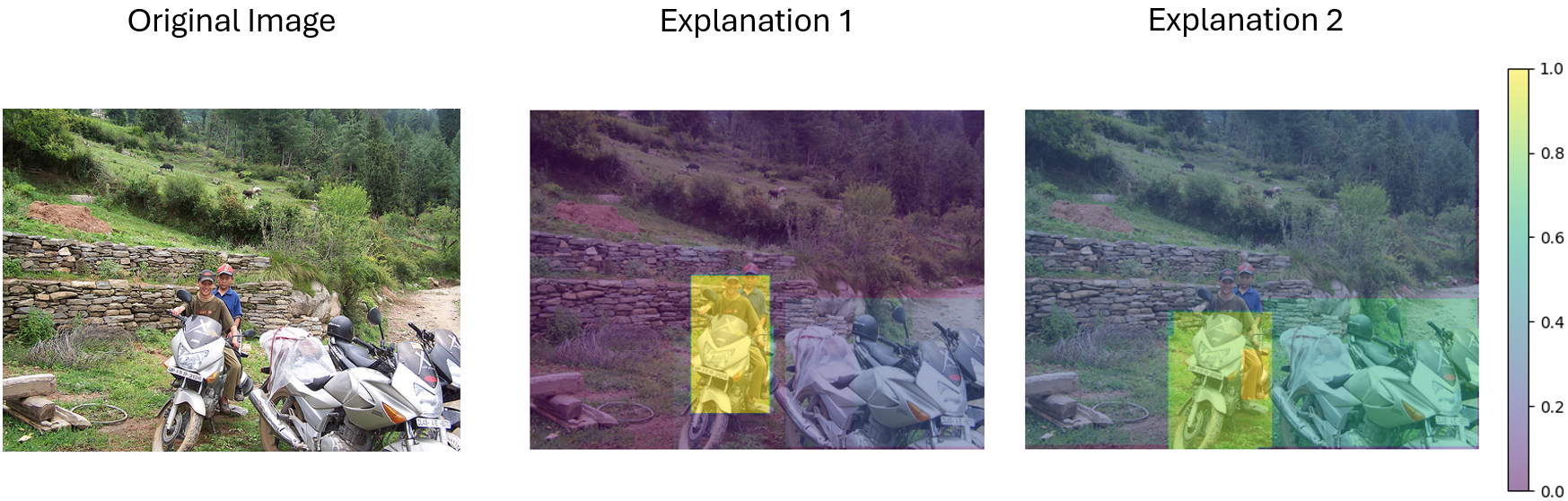}} & \parbox[t]{0.4\textwidth}{\vspace{-50pt}{\fontsize{7}{8}\selectfont Q: Which explanation is for object "person" and which is for object "motorbike"? \\ A1: Explanation 1 explains "person" and Explanation 2 explains "motorbike" \\ A2: Explanation 1 explains "motorbike" and Explanation 2 explains "person"\\ Correct answer is A1 with $100\%$ correct answers.}}\vspace{10pt}\\
    \parbox[t]{0.4\textwidth}{\includegraphics[width=0.5\textwidth]{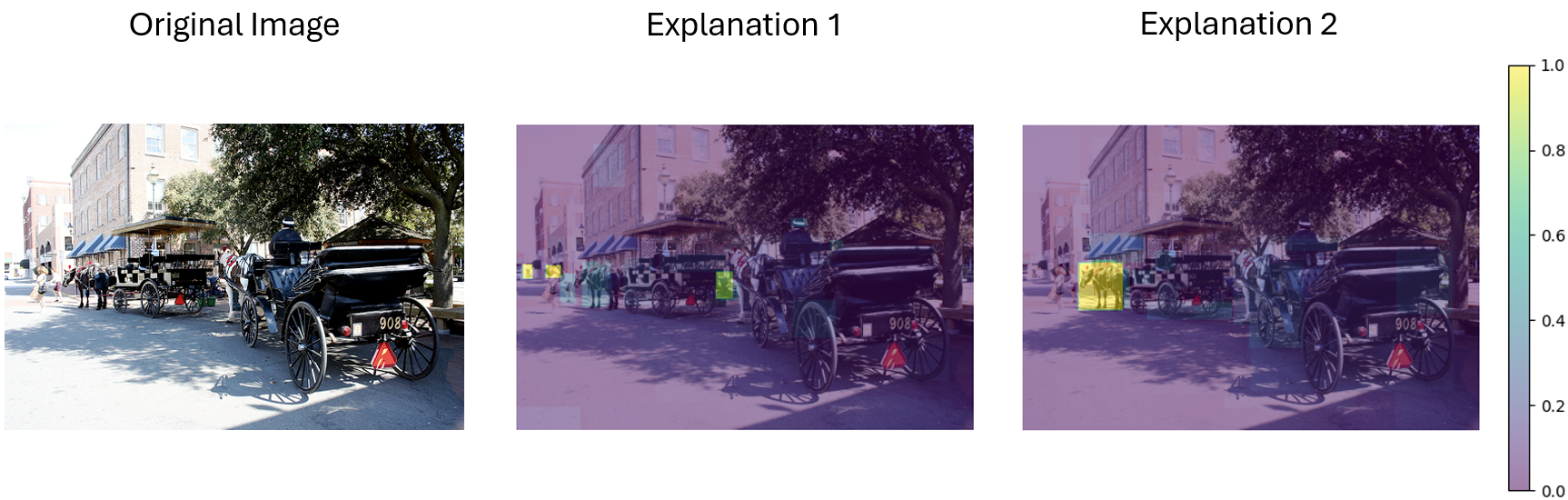}} & \parbox[t]{0.4\textwidth}{\vspace{-50pt}{\fontsize{7}{8}\selectfont Q: Which explanation is for object "person" and which is for object "horse"? \\ A1: Explanation 1 explains "person" and Explanation 2 explains "horse" \\ A2: Explanation 1 explains "horse" and Explanation 2 explains "person" \\ Correct answer is A1 with $100\%$ correct answers.}}\vspace{10pt}\\
\end{longtable}

\FloatBarrier

\section{Bias Assessment}
As detailed in Section 4.4, we employ Stable Diffusion XL (SDXL) to generate images and construct the portrait datasets used to assess model biases. We discuss generation details and give examples of the generated images.
For the \textit{biological gender} bias, we generate $50$ portraits of female and $50$ portraits of male people using prompt:
\begin{quote}
    ``\texttt{A headshot of a {\color{blue} <man|woman>} in his 40s, DSLR, detailed, 8k, in perfect focus, shoulder height}''
\end{quote}
For the \textit{ethnicity} bias, we generate $50$ portraits of african-american and $50$ portraits of caucasian people using prompt:
\begin{quote}
    ``\texttt{A headshot of a {\color{blue} <caucasian|african-american>} \,person, DSLR, detailed, 8k, in perfect focus, shoulder height}''
\end{quote}
We use Stability AI's official SDXL checkpoint \texttt{stable-diffusion-xl-base-1.0}, and run it using a diffuse-refine pipeline with $40$ initial diffusion steps and $10$ refinement steps. All generated images are publicly available at \texttt{[redacted for anonimity]}.
\vspace{40pt}

\begin{figure}[h]
\centering
\begin{tabular}{cc}
\includegraphics[width=0.45\linewidth]{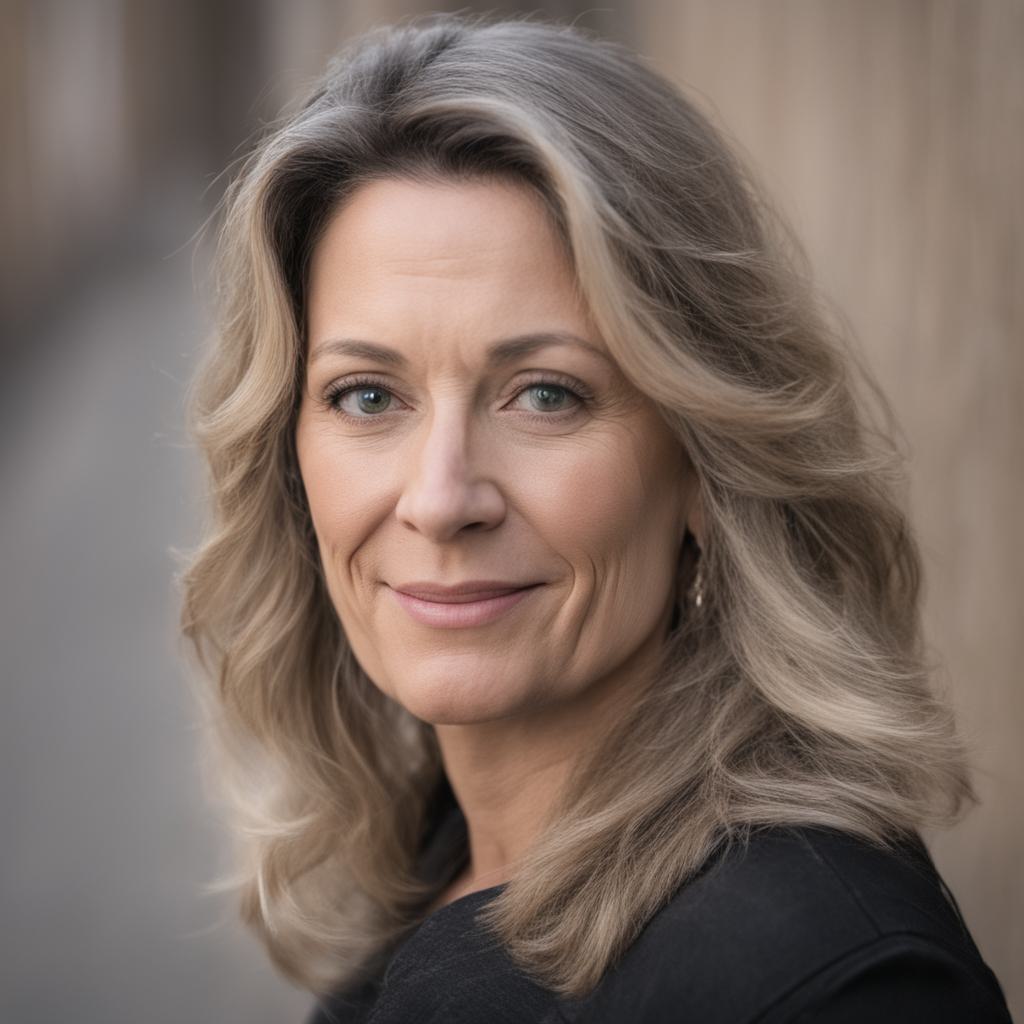} & \includegraphics[width=0.45\linewidth]{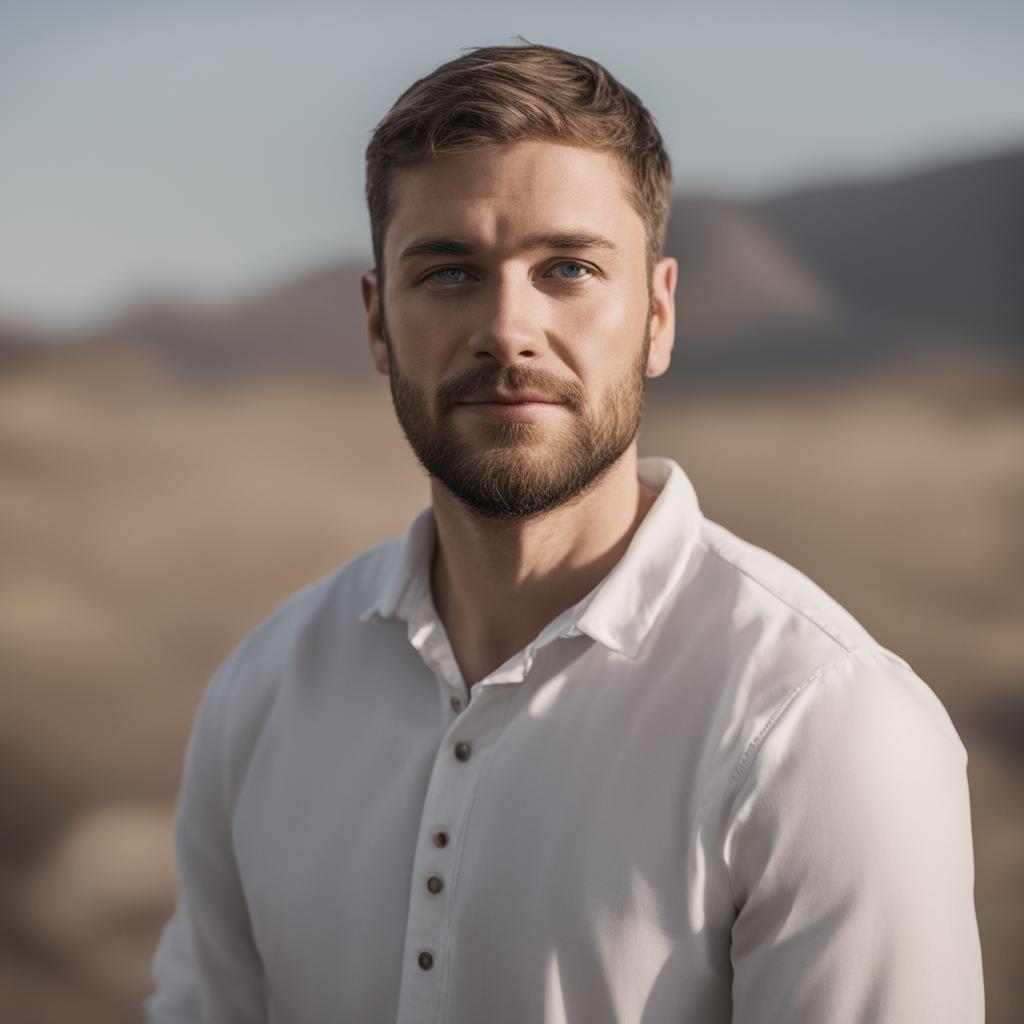} \\
female person & male person \vspace{10pt}\\
\includegraphics[width=0.45\linewidth]{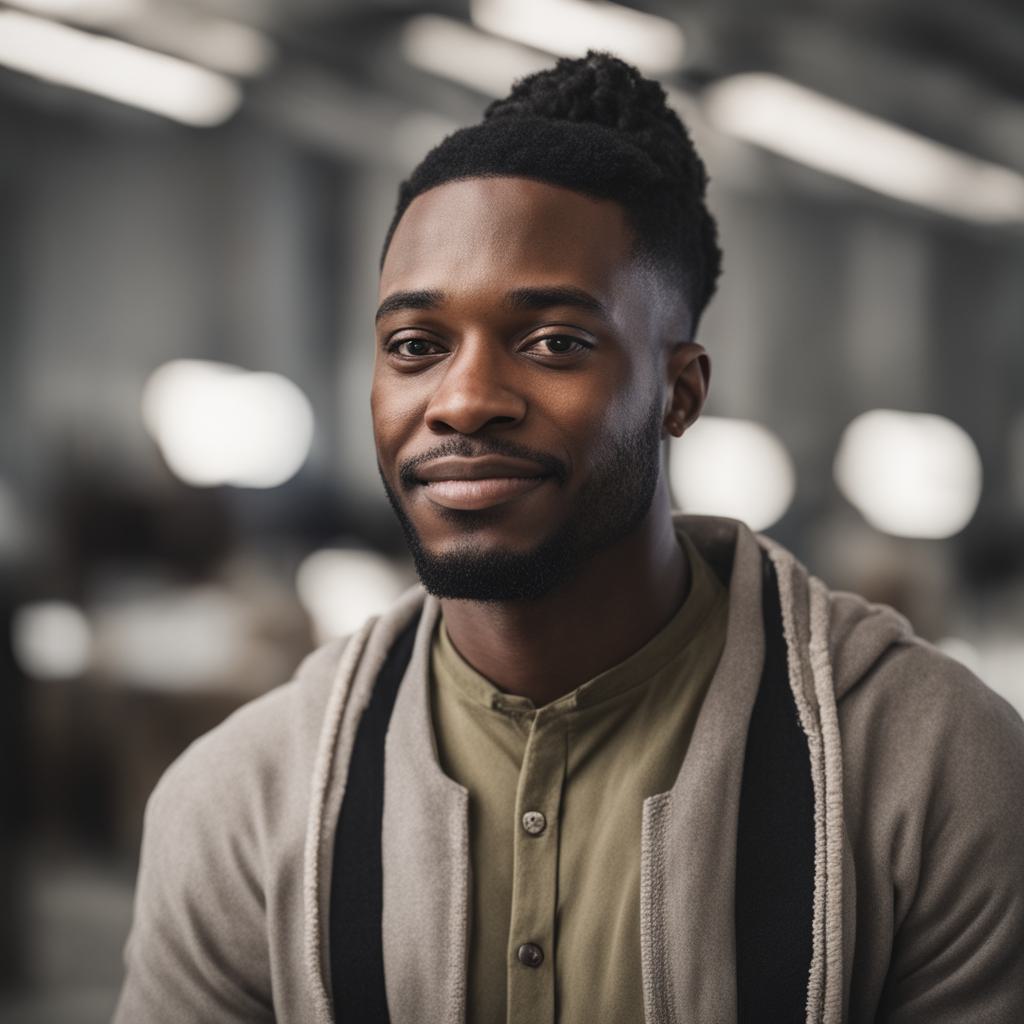} & \includegraphics[width=0.45\linewidth]{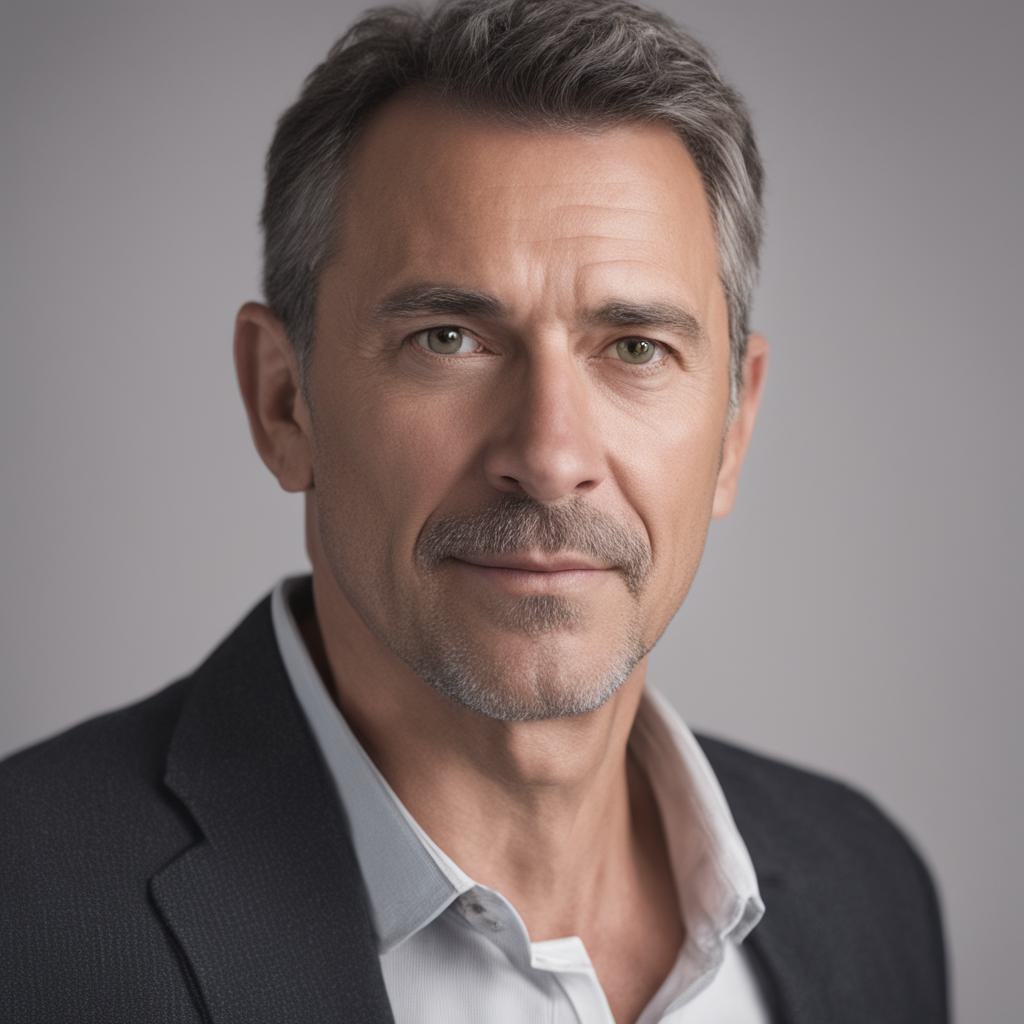} \\
African-American person & Caucasian person \vspace{10pt}\\
\end{tabular}
\caption{Example images generated using each of the four prompts.}
\end{figure}

\bibliography{egbib}